\documentclass[10pt,journal,compsoc]{IEEEtran}

\usepackage{url}            % simple URL typesetting
\usepackage{booktabs}       % professional-quality tables
\usepackage{amsfonts}       % blackboard math symbols
\usepackage{nicefrac}       % compact symbols for 1/2, etc.
\usepackage{microtype}      % microtypography
\usepackage{xcolor}         % colors
\usepackage{amsmath}
\usepackage{graphicx}
\usepackage{algorithm}
\usepackage{algorithmic}
\usepackage{multicol}
\usepackage{multirow}
\usepackage{stfloats}
\makeatletter

\newcommand{\Rmnum}[1]{\expandafter\@slowromancap\romannumeral #1@}
\makeatother
\newcommand{\etal}{{\emph{et al.}}}
\newcommand{\ie}{{\emph{i.e.}}}
\newcommand{\eg}{{\emph{e.g.}}}

\usepackage{bm}
\usepackage{epigraph}

\ifCLASSOPTIONcompsoc
  \usepackage[nocompress]{cite}
\else
  \usepackage{cite}
\fi

\ifCLASSINFOpdf
\else
\fi
\usepackage[colorlinks,linkcolor=red,citecolor=blue]{hyperref}       % hyperlinks
\usepackage{url}

\begin{document}
\title{Towards Context-Aware Emotion Recognition Debiasing from a Causal Demystification Perspective via De-confounded Training}

\author{Dingkang Yang, Kun Yang, Haopeng Kuang, Zhaoyu Chen, Yuzheng Wang, Lihua Zhang,~\IEEEmembership{Member, IEEE}
% \\ Song Wang,~\IEEEmembership{Senior Member, IEEE},
\IEEEcompsocitemizethanks{\IEEEcompsocthanksitem Dingkang Yang, Kun Yang, Haopeng Kuang, Zhaoyu Chen, Yuzheng Wang, and Lihua Zhang are with the Academy for Engineering and Technology, Fudan University, Shanghai 200433, China. (E-mail: \{dkyang20, kunyang20, hpkuang19, zhaoyuchen20, yzwang20, lihuazhang\}@fudan.edu.cn). Corresponding author: Lihua Zhang.
\IEEEcompsocthanksitem
This work is supported in part by the National Key R\&D Program of China (No.\,2021ZD0113503) and in part by the Shanghai Municipal Science and Technology Major Project (No.\,2021SHZDZX0103).
}}

\markboth{IEEE TRANSACTIONS ON PATTERN ANALYSIS AND MACHINE INTELLIGENCE, 2024}%
% \markboth{Journal of \LaTeX\ Class Files,~Vol.~14, No.~8, August~2015}%
{Towards Context-Aware Emotion Recognition Debiasing from a Causal Demystification
Perspective via De-confounded Training}

\IEEEtitleabstractindextext{%
\begin{abstract}

Understanding emotions from diverse contexts has received widespread attention in computer vision communities. The core philosophy of Context-Aware Emotion Recognition (CAER) is to provide valuable semantic cues for recognizing the emotions of target persons by leveraging rich contextual information. Current approaches invariably focus on designing sophisticated structures to extract perceptually critical representations from contexts. Nevertheless, a long-neglected dilemma is that a severe context bias in existing datasets results in an unbalanced distribution of emotional states among different contexts, causing biased visual representation learning. From a causal demystification perspective, the harmful bias is identified as a confounder that misleads existing models to learn spurious correlations based on likelihood estimation, limiting the models' performance. To address the issue, we embrace causal inference to disentangle the models from the impact of such bias, and formulate the causalities among variables in the CAER task via a customized causal graph. Subsequently, we present a Contextual Causal Intervention Module (CCIM) to de-confound the confounder, which is built upon backdoor adjustment theory to facilitate seeking approximate causal effects during model training. As a plug-and-play component, CCIM can easily integrate with existing approaches and bring significant improvements. 
Systematic experiments on three datasets demonstrate the effectiveness of our CCIM.

\end{abstract}

}

\maketitle

\IEEEdisplaynontitleabstractindextext

\IEEEpeerreviewmaketitle

\IEEEraisesectionheading{\section{Introduction}\label{sec:intro}}

\setlength{\epigraphwidth}{0.45\textwidth} 
\epigraph{\emph{``Context is the key to understanding, but it can also be the key to misunderstanding.''}}{\scriptsize-\emph{Jonathan Lockwood Huie}}

\begin{figure}[t]
  \centering
  \includegraphics[width=0.92\linewidth]{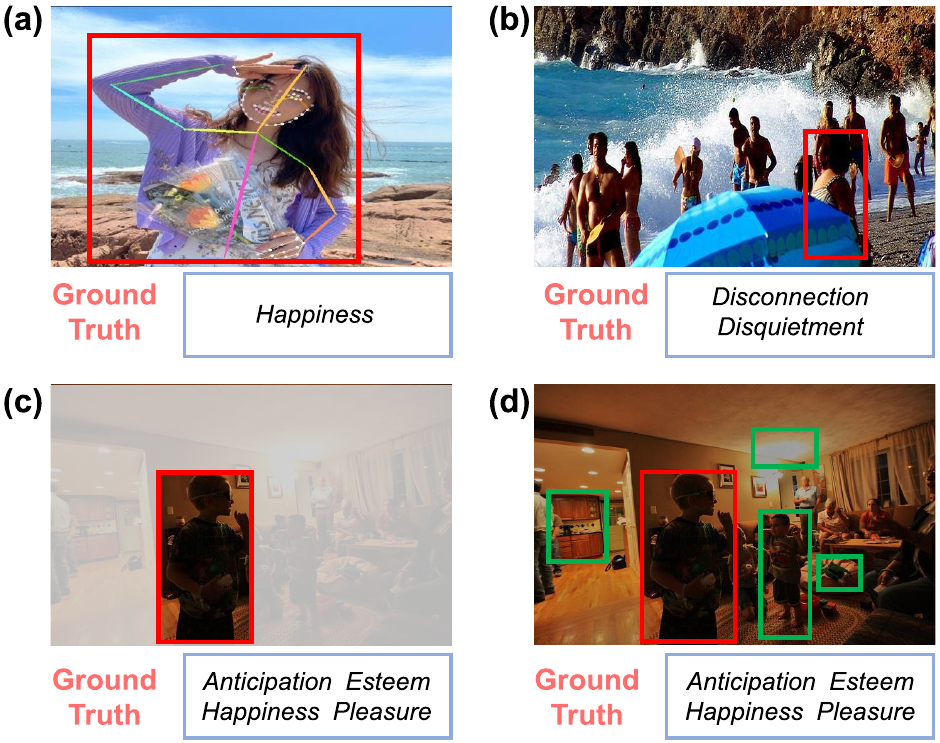}
  \caption{We provide several examples of emotion recognition in non-controlled scenarios. The red bounding boxes include the recognized subjects. (a) shows the ideal case of subject-centered emotion recognition, where previous efforts have extracted emotion-related semantics from available face, posture, and gesture information. (b) shows the common dilemma in the wild environment, where the subject's bodily regions are usually indistinguishable. In (c), it is difficult to recognize the emotion of the vague subject where the surrounding context is obscured. (d) shows complementary cues from the visible context around the subject that may reflect emotion, which is localized by the green bounding boxes.
  }
  \label{indro}
\vspace{-0.3cm}
\end{figure}

\IEEEPARstart{A}{s} an essential element of human experience, emotion significantly influences social interactions and communication~\cite{kossaifi2019sewa,wagner2023dawn}. Accurately identifying human emotions from visually accessible content has become integral to pattern recognition methods~\cite{liu2023brain}.
Due to the promising application prospects in understanding human intentions and expressions, emotion recognition has received widespread attention in various fields, such as assisted driver monitoring~\cite{yang2023aide}, online education~\cite{tanko2022shoelace}, and human-machine interaction~\cite{alnuaim2022human}. For instance, intelligent vehicle systems can automatically detect drivers' emotional states and provide the necessary alerts to reduce road safety hazards when subjects are distracted~\cite{yang2023aide}.

Conventional emotion recognition in computer vision mainly focuses on subject-centered representation channels, including but not limited to facial expressions~\cite{du2021learning,farzaneh2021facial}, bodily postures~\cite{wallbott1998bodily,mahfoudi2022emotion}, skeletal gestures~\cite{liu2021imigue,shen2019emotion}, and multimodal combinations~\cite{yang2022disentangled,yang2023target,yang2022learning,yang2022contextual}. Figure \ref{indro}(a) presents an intuitive example of these endogenous factors from facial landmarks and body keypoints, providing valuable multi-source cues to recognize the subject's happiness.
Despite the remarkable advancements,
their performance suffers the inevitable dilemma in uncontrolled wild scenarios. As shown in Figure \ref{indro}(b), the perceptually critical regions of the subject in field-collected data are generally indistinguishable (\eg, ambiguous gestures) due to natural occlusions or recorded viewpoints. In this case, the representation channels from the subject would fail to provide meaningful emotional signals.

Recently, several emerging approaches~\cite{kosti2017emotion,lee2019context,zhang2019context,mittal2020emoticon,de2023high} have suggested capturing additional emotion semantics from out-of-subject contexts to overcome performance bottlenecks in real-world applications. According to the pioneering work~\cite{kosti2019context}, contexts are considered to include diverse surrounding factors, such as place attributes, scene concepts, background objects, and the actions of others nearby the subject. Cognitive psychology research~\cite{barrett2011context} has demonstrated that different contexts spontaneously affect human emotional states in society and offer complementary affective cues. An interesting illustration is given in Figure \ref{indro}(c)\&(d). When the situational context is ignored in Figure \ref{indro}(c), we can just observe a blurred outline of the subject, and hard to recognize the probable emotion polarity. In contrast, we find the subject interacting with his family or friends in a warm room with a pleasurable atmosphere when the context is visible in Figure \ref{indro}(d). Although physical signals are ambiguous, exogenous stimuli from the surrounding context can help us better infer the positive state of the subject that he may feel anticipated, esteemed, or pleased. This promising technology for combining contextual information is called Context-Aware Emotion Recognition (CAER).

\begin{figure}[t]
  \centering
  \includegraphics[width=\linewidth]{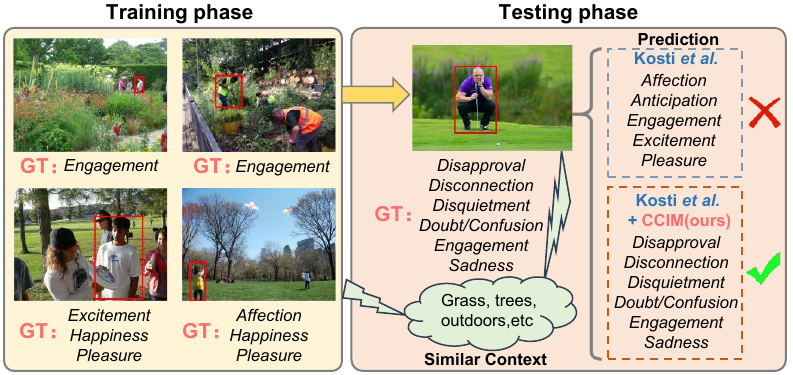}
  \caption{The harmful context bias in the CAER task is intuitively demonstrated by randomly selecting sample examples in the training and testing sets of the EMOTIC dataset. GT represents the ground truth of samples. Most training samples containing vegetated surround contexts have similar positive emotion categories. In this case, the model~\cite{kosti2019context} relies on spurious correlations between specific contexts and emotion categories to learn misleading visual representations, causing entirely incorrect predictions. Thanks to the proposed CCIM, the model automatically corrects the prediction errors and gives more accurate results.
  }
  \label{indro1}
\vspace{-0.3cm}
\end{figure}

Current mainstream CAER studies usually follow a common procedure pipeline: (1) extracting endogenous characteristics from the recognized subject's region; (2) exploring different context branches and learning emotion-related representations; (3) 
constructing well-designed fusion mechanisms to integrate these features for downstream emotion label predictions.
Despite the considerable improvements achieved by existing methods relying on sophisticated model structures~\cite{mittal2020emoticon,zhang2019context,ruan2020context,li2021sequential,gao2021graph,chen2023incorporating} or different fusion strategies~\cite{mittal2021multimodal,hoang2021context,yang2022emotion,wang2022context}, they invariably suffer from a context bias of the datasets. To our best knowledge, it is the first time that this long-neglected problem has been investigated. Reflecting on the process of creating CAER datasets, varied annotators were tasked with labeling each image based on their subjective thoughts of the emotions being experienced by individuals in the images across a range of contexts~\cite{kosti2019context}. This procedure unavoidably influences the distribution of emotion categories across various 
contexts due to annotators' preferences, consequently resulting in the context bias. Figure~\ref{indro1} provides an intuitive demonstration of how understanding such bias confounds emotion predictions.
The training data primarily encompasses images showcasing scenes abundant in vegetation, which are associated with positive emotion categories. Conversely, instances of negative emotions within analogous contexts are notably rare.
As a consequence, the baseline model~\cite{kosti2019context} has the potential to be misguided, acquiring spurious correlations between context-specific features and label semantics. When confronted with the testing image featuring similar contexts yet containing negative emotion categories, the model inevitably arrives at inaccurate deductions about emotional states.

 \begin{figure}[t]
  \centering
  \includegraphics[width=\linewidth]{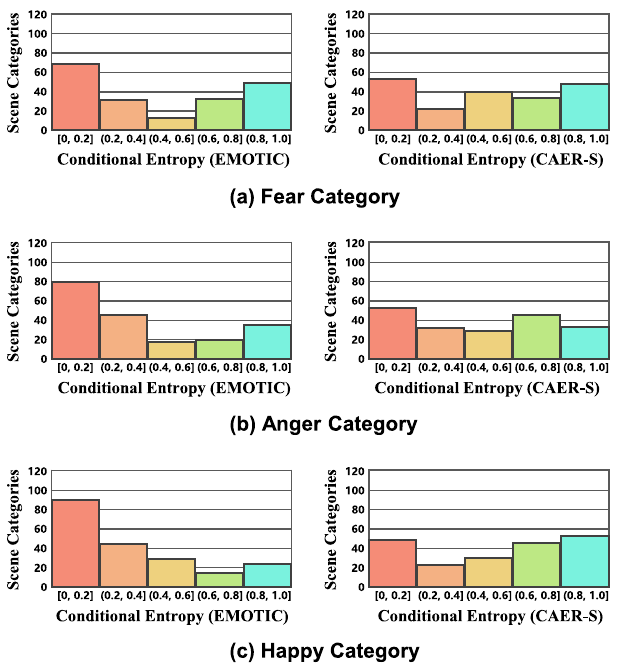}
  \caption{We present a preliminary toy experiment using the EMOTIC~\cite{kosti2019context} and CAER-S~\cite{lee2019context} datasets, focusing on scene categories associated with fear, anger, and happy emotions. The inclusion of more scene categories exhibiting normalized zero-conditional entropy reveals a pronounced presence of the harmful context bias.
  }
  \label{indro2}
\vspace{-0.3cm}
\end{figure}

More intrigued, we perform a toy experiment on two CAER datasets to further verify the severe context bias through the scene context attributes. This preliminary test focuses on quantitatively investigating how emotions are related to contexts (\eg, scene categories). Concretely speaking, we use the ResNet-152~\cite{he2016deep} pre-trained on the Places365 dataset~\cite{zhou2017places} to predict scene categories from images with three common emotion categories (\ie, ``fear'', ``anger'', and ``happy'') across two datasets.
The selection process involves identifying the 200 scenes with the highest occurrence rates in each emotion category. Subsequently, the normalized conditional entropy is calculated across positive and negative subsets of a specific emotion~\cite{panda2018contemplating}. Given a scene category $c$, the conditional entropy
is computed as $\mathcal{H}(Y|X=c)=-\sum_{y \in \{ e_p,e_n \} }^{} p(y|X=c) {\rm log} p(y|X=c)$, where $e_p$ and $e_n$ denote the positive and negative subsets of emotion $e$ respectively 
(\eg, ``happy'' and ``non-happy'').
While examining associations between scene contexts and emotion categories in Figure \ref{indro2}, it becomes evident that more occurrence of scene categories featuring zero conditional entropy likely implies the existence of the notable context bias within the datasets. This is characterized by scenes exclusively appearing either in the positive or negative subsets of emotions.
Specifically, 33\% and 27\% of the scene categories targeting fear on the EMOTIC and CAER-S datasets, respectively, are in the entropy range of [0, 0.2].
Within the EMOTIC dataset~\cite{kosti2019context}, approximately 40\% of anger-related scene categories exhibit zero conditional entropy, while around 45\% of the categories for happy (\ie, happiness) have zero conditional entropy. As a tangible example, scene contexts closely related to celebrations are predominantly present in instances with the happy category, while their presence is virtually absent in the negative emotion categories. These findings substantiate the pronounced context bias within the CAER datasets, resulting in discernible disparities in the distribution of emotion categories across various contexts and imbalanced visual representations.

Motivated by the above analyses, we attempt to embrace causal inference~\cite{pearl2009causal} to reveal the culprit that poisons the CAER models, rather than focusing on beating the previous approaches. 
As a groundbreaking scientific paradigm that propels models towards unbiased predictions, the primary hurdle in applying traditional causal inference to the contemporary CAER task lies in effectively depicting genuine causal effects and recognizing task-specific dataset bias. To this purpose, we seek causalities rather than shallow associations to improve bias-plagued models from a causal demystification perspective~\cite{pearl2009causal}. Concretely, we propose a causality-based bias mitigation component that is simple in implementation but powerful in functionality. A tailored structured causal model is first presented to explain the causal procedure of the CAER task.
In this case, the harmful \textbf{context bias} in datasets is essentially an unintended \textbf{confounder} that misleads the models to learn the spurious correlation between similar contexts and specific emotion semantics.
We decouple the causal dependencies among the input images $\bm{X}$, subject features $\bm{S}$, context features $\bm{C}$, confounder $\bm{Z}$, and predictions $\bm{Y}$.
Essentially differentiating from conventional likelihood estimation $P(\bm{Y}|\bm{X})$, we propose a Contextual Causal Intervention Module (CCIM) to accomplish context de-confounding during model training with a novel $do$-operation $P(\bm{Y}|do(\bm{X}))$.
As a causal intervention philosophy, $do(\cdot)$ operator can effectively prevent the establishment of spurious correlations among variables in the non-causal direction.
CCIM is based on the backdoor adjustment theory~\cite{glymour2016causal} to approximate true causal effects and remove the unfavorable impact of the confounder caused by the context bias. As a plug-and-play, model-agnostic, and lightweight component, CCIM can be easily integrated into existing baselines and bring significant and consistent improvements.
Quantitative and qualitative experiments demonstrate the necessity and effectiveness of the proposed CCIM. 
Our main contributions follow below:

\begin{itemize}
\item To our best knowledge, we are the first to disentangle variables in the CAER task from the causal demystification perspective and to deeply investigate deleterious context bias in the datasets. Such bias is essentially an unplanned confounder to mislead CAER models to unconsciously capture spurious correlations and misinterpret context semantics.

\item We present a Contextual Causal Intervention Module (CCIM) through theoretical derivations based on the backdoor adjustment and the practical implementation based on network parameterization. CCIM can be incorporated into most CAER models to achieve a fair contribution of different contexts to emotion understanding by approximating true causal effects.

\item Extensive experiments are implemented on three standard CAER datasets. Systematic analyses show the potential of the proposed CCIM to improve existing models and thus enable bias-free predictions.
\end{itemize}

This work significantly extends our preliminary paper~\cite{yang2023context} at the \textit{IEEE International Conference on Computer Vision and Pattern Recognition (CVPR2023)}.
We provide improvements in multiple aspects to enhance further the applicability and scalability of our work.
Specifically, \textbf{(i)} we introduce the emotional state model based on continuous dimensions to assess and show the context bias dilemma. The continuous emotion representations contain three subspaces, \textit{Valence}, \textit{Arousal}, and \textit{Dominance}, contributing to a complementary evaluation of CCIM's gain in complex emotion-binding situations;
\textbf{(ii)} we present the Jaccard coefficient scores to more abundantly explain the different performances and roles of our causal intervention in diverse context instances. These criteria give intuitive explanations for measuring the differences between the causal intervention process and the traditional likelihood estimation procedure; \textbf{(iii)} considering that a subject may have multiple emotion intentions in multi-label emotion recognition, label-based and sample-based evaluation rules are proposed to measure existing methods and provide additional prediction results more rationally; \textbf{(iv)} we combine CCIM with more state-of-the-art (SOTA) approaches. The noteworthy performance improvements of CCIM on model structures with distinct design philosophies and fusion mechanisms enhance the persuasiveness and usefulness of our work; \textbf{(v)} more details and discussions on the motivation, algorithms, and implementation are provided to reinforce our insights. Furthermore, more experiments are conducted to highlight the effectiveness of CCIM, including quantitative, qualitative, ablative, and customized analyses.

The rest of this paper is organized as follows. In Section~\ref{related}, we discuss the background among related techniques in prior works, including uni/multimodal emotion recognition, context-aware emotion recognition, and causal demystification. The detailed methodology is provided in Section~\ref{method}, which consists of the causal graph construction, the causal intervention interpretation, and the parameterized implementation. Section~\ref{dataset} introduces used standard datasets, representative models, and evaluation metrics. The systematic experiments and analyses are described in Section~\ref{detail}. Finally, our conclusions are drawn in Section~\ref{con}.

\section{Related Work}
\label{related}
\subsection{Uni/Multimodal Emotion Recognition}
Emotion is an essential medium for humans to communicate their intentions and maintain social relationships with the external world~\cite{barrett2011context}. As an important component in the affective computing fields, emotion recognition technology has received widespread attention and exploration over the past decade~\cite{kossaifi2019sewa,wagner2023dawn,liu2023brain,yang2023aide,tanko2022shoelace}.
As a complex psychological activity, emotion descriptions are generally summarized in two directions: discrete categories and continuous dimensions. The basic discrete emotions~\cite{ekman1971constants} are categorized as Happiness, Fear, Surprise, Sadness, Disgust, and Anger.
Several subsequent taxonomies~\cite{yang2023aide,li2020deep} are combinations or refinements of these six typical emotions.
Continuous dimensions usually utilize numerical representations of sequential descriptors to granularly depict different emotion subspaces. The most influential way is the VAD emotion model~\cite{mehrabian1995framework}, which decouples emotional states into three dimensions regarding Valence, Arousal, and Dominance.
Early works focused on unimodal recognition patterns, which were dominated by facial expression analysis.
The face-oriented approaches~\cite{du2021learning,farzaneh2021facial,pantic2000expert} usually attend to geometric or appearance characteristics for extracting informative representations that reflect emotions.
Another research direction~\cite{wallbott1998bodily,mahfoudi2022emotion,liu2021imigue,shen2019emotion,schindler2008recognizing} is to explore the emotional semantics embedded in body language through gestural or postural information.
More recently, multimodal emotion recognition aims to aggregate heterogeneous modalities from different channels to jointly learn emotion-related representations~\cite{yang2022disentangled,yang2023target,yang2022learning,yang2022contextual,lei2023text,li2024unified,li2023towards}.
Despite impressive advances, previous efforts have generally been restricted to laboratory-controlled or well-designed settings.
The performance of subject-centered approaches may deteriorate when modalities are missing or signals are ambiguous in uncontrolled real-world scenarios.

\subsection{Context-Aware Emotion Recognition}
Context-Aware Emotion Recognition (CAER)~\cite{kosti2017emotion} provides new possibilities for robust affective interactions in uncontrolled wilderness environments. 
As an emerging task, CAER not only incorporates the subject-centered learning paradigm, but also considers substantial emotion cues from out-of-subject contexts.
Existing approaches~\cite{kosti2019context,lee2019context,mittal2020emoticon,zhang2019context,wang2022context,li2021sequential,li2021human,hoang2021context,gao2021graph,ruan2020context,yang2022emotion,chen2023incorporating} typically extract multiple representations from subject and context sources to perform feature fusion and subsequent label prediction.
Specifically, Kosti \etal \cite{kosti2017emotion} first treat the complete image as a global context support and implement a two-stream Convolutional Neural Network (CNN) model with low-rank filtering properties.
Zhang \etal \cite{zhang2019context} utilize the region proposal network~\cite{ren2015faster} to select different elements in the background context to construct an affective graph and infer emotional states. Then, CAGBN~\cite{ruan2020context} is proposed to fuse global and local information in the images with the view of a sequence generation task.
Besides using multiple modalities from the subject,
EmotiCon \cite{mittal2020emoticon} introduces the scene and socio-dynamic contexts following Frege’s Principle.
After that, SIB-Net~\cite{li2021sequential} is presented to capture the relation of sequence and interaction among face, body, and scene context. 
In the recent methodology, Yang \etal \cite{yang2022emotion} discover the fine-grained relationship between agents and objects to mitigate the uncertainty of contextual semantics in expressing emotions from a sociological perspective.
While previous approaches have achieved promising improvements by seeking rich complementary factors from diverse contexts, they have all ignored the intrinsic dilemma of performance bottlenecks caused by the context bias of the datasets.
Instead of focusing on beating the latest SOTA, we step back to disentangle the harmful bias from a novel causal perspective and bring consistent gains for existing models via the proposed CCIM.

\subsection{Causal Demystification}
Causal demystification is an essential application of causal inference, which aims to analyze the intrinsic dynamics and possible outcomes of events when the corresponding conditional variables are changed~\cite{pearl2009causal}.
By adjusting different treatments and interventions, this theory has been widely applied and achieved considerable exploration in various fields, such as economics~\cite{varian2016causal} and developmental psychology~\cite{foster2010causal}. 
Maintaining the principle of universal applicability, the pursuit of causal demystification bifurcates into two fundamental avenues: the structured causal model~\cite{pearl2000models} and the potential outcome framework~\cite{rubin2005causal}. These dual methodologies serve as illuminative instruments, delving into the underpinnings of causalities rather than remaining confined to the realm of superficial variable associations.
Some early works attempt to provide reliable explanations for the models by relying on causal theories~\cite{pearl2014interpretation,kocaoglu2017causalgan}.
Benefiting from learning-based technologies~\cite{yang2023how2comm,yang2023spatio,yang2023what2comm,yang2024TSIF,MRG3Net10419057,chen2024can,chen2024detecting,jiang2024medthink,chen2024efficiency}, modern deep learning tasks~\cite{yang2024SuCi,yang2024MCIS,yang2024robust} have begun to embrace causal tools for unbiased estimation solutions, including computer vision~\cite{wang2022spacenet,wang2020visual,qi2020two,tang2020unbiased,chen2022causal} and natural language processing~\cite{zhang2021biasing,qian2021counterfactual,huang2020counterfactually}.
Unlike the task-specific causal paradigm described above, this is the first investigation of the confounding effect through causal inference in the CAER task while exploiting causal intervention to interpret and address the confounding bias from contexts.

 \begin{figure}[t]
  \centering
  \includegraphics[width=\linewidth]{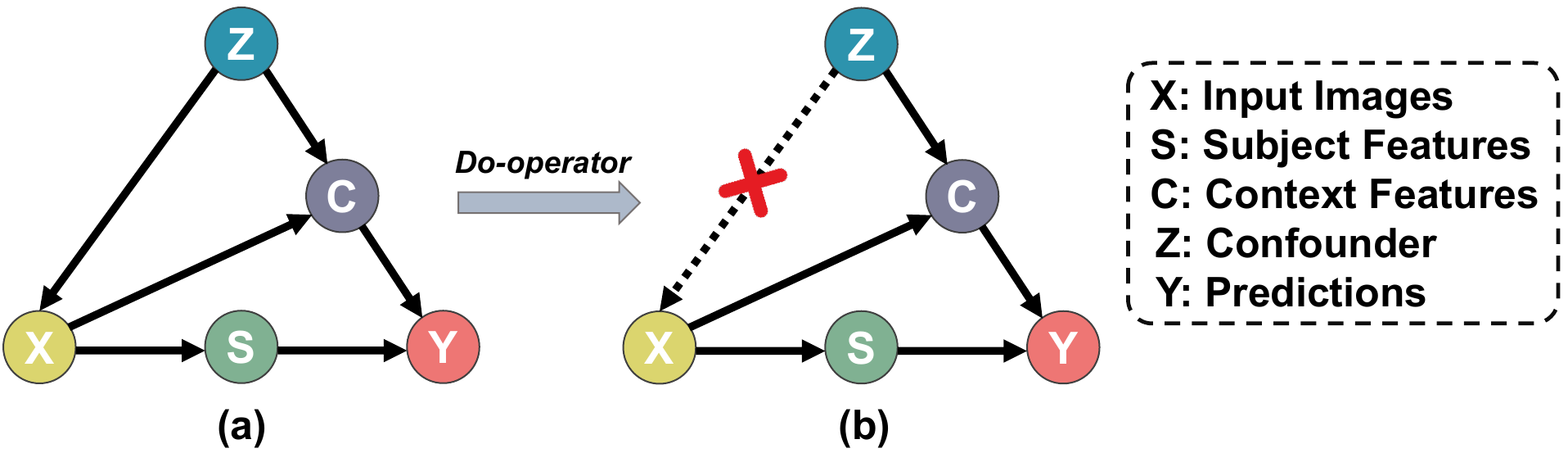}
  \caption{Illustration of our CAER causal graph. (a) The conventional likelihood $P(\bm{Y}|\bm{X})$. (b) The causal
intervention $P(\bm{Y}|do(\bm{X}))$.
  }
  \label{indro3}
\vspace{-0.3cm}
\end{figure}

\section{Methodology}
\label{method}
\subsection{Causal Perspective at CAER Task}
Before starting, we present a customized causal graph to disentangle the general CAER process. In particular, we adhere to the same graphical notation within the framework of the structured causal model~\cite{pearl2000models}, attributing this choice to its inherent quality of intuitive lucidity and facilitative interpretability. 
The causal graph is formally a directed acyclic graph $\mathcal{G} = \{ \mathcal{N}, \mathcal{E} \}$ that can be utilized to achieve causal estimates across data. The nodes $\mathcal{N}$ denote variables and the links $\mathcal{E}$ denote direct causal effects. As illustrated in Figure~\ref{indro3}, the CAER procedure contains five different variables, which are input images $\bm{X}$, subject features $\bm{S}$, context features $\bm{C}$, confounder $\bm{Z}$, and predictions $\bm{Y}$.
Note that the proposed causal graph is well adapted to a wide range of CAER models because it is highly summarized and not restricted by implementation details. The comprehensive exposition regarding the underlying architecture of these causal interconnections is furnished hereinafter.

\textbf{Link $\bm{Z} \rightarrow \bm{X}$}.
In uncontrolled environments, distinct subjects are recorded by the publishers of the datasets in diverse context scenarios to produce image samples $\bm{X}$. When assessing emotional states the subjects evinced, the annotators generally provide possible emotion annotations with biased and subjective awareness~\cite{kosti2019context,lee2019context}.
Despite adopting several qualifications and control measures for annotators, bounded human observations of the natural world inevitably lead to biased annotation performance~\cite{jones1999bounded}. From the intuitive example in Figure~\ref{indro1}, subjects are habitually ascribed positive emotion categories within contexts rich in vegetative cover, with such assignments often transpiring without overt cognitive deliberation.
Another nonnegligible reason is that the data nature results in the unbalanced distribution of the emotional states in the real world~\cite{geirhos2020shortcut}. Fundamentally, the collection of positive emotions in contexts characterized by comfort is markedly less challenging than in those marked by negativity.
The context bias induced by the above situations is identified as a harmful confounder $\bm{Z}$, which establishes spurious associations between similar context representations and specific emotion semantics.
To be precise, the confounder $\bm{Z}$ directly determines the recorded biased content in the input images $\bm{X}$, \ie, $ \bm{Z} \rightarrow \bm{X}$.

\textbf{ Link $\bm{Z} \rightarrow \bm{C} \rightarrow \bm{Y} $.}
$\bm{C}$ implies that the total context features come from the context representation encoders. Since the node variable $\bm{C}$ is a generalized descriptor, its specific implementation depends on the definition and modeling of contexts by different methods. For instance, context features may derive from the background region after hiding the subject's face~\cite{lee2019context} or from the aggregation of scene and socio-dynamic context information~\cite{mittal2020emoticon}. The causal path $ \bm{Z} \rightarrow \bm{C} $ represents the deleterious confounder $\bm{Z}$ misleading the models to capture the contextual semantics from $\bm{C}$, which has unreliable emotion correlations. In this situation, the unpure $\bm{C}$ would further impact the predictions of the emotion labels, and its effect would propagate along the link $\bm{C} \rightarrow \bm{Y}$. A noteworthy point is that $\bm{Z}$ potentially contains prior knowledge from the training data to assist the models in estimating appropriately when the subject's characteristics are indistinguishable. Nevertheless, the confounding attributes in $\bm{C}$ largely mislead the models to learn spurious ``\emph{context-emotion}" mapping during training, causing biased predictions with performance bottlenecks.

 \begin{figure*}[t]
  \centering
  \includegraphics[width=0.9\linewidth]{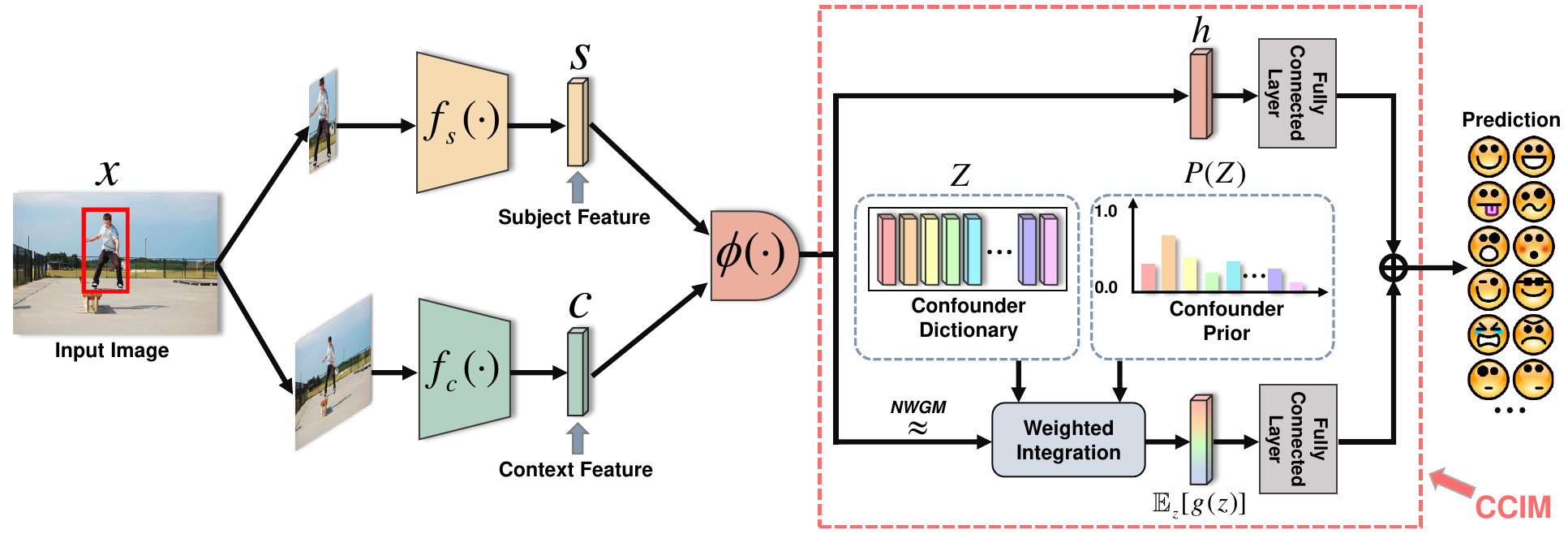}
  \caption{We present a general pipeline for the context-deconfounded training in the CAER task. The pipeline can be adapted to most CAER models. Given an input image $\bm{x}$, two generalized coding functions $f_{s} (\cdot)$ and $f_{c} (\cdot)$ extract the subject feature $\bm{s}$ and context feature $\bm{c}$ from different regions, respectively. Subsequently, $\bm{s}$ and $\bm{c}$ are integrated and obtain the joint representation $\bm{h}$ through a fusion strategy whose specific implementation follows different methods. The red dotted box shows the core component: the proposed CCIM. Our CCIM is inserted before the task-specific classifier to reasonably approximate the causal intervention and assist the models in seeking the true causal effect during training.
  }
  \label{intervention}
\end{figure*}

\textbf{Link $\bm{X} \rightarrow \bm{C}
\rightarrow \bm{Y} $ \& $ \bm{X} \rightarrow \bm{S}
\rightarrow \bm{Y} $.}
$\bm{S}$ represents the total subject features obtained by subject representation encoders. Similarly to $\bm{C}$, the detailed implementation of $\bm{S}$ is also not limited to a specific method. That is, subject features could be extracted from appearance characteristics of the body region~\cite{li2021human} or the cropped face position~\cite{lee2019context}. In the CAER causal graph, the desperately desired effects of input images $\bm{X}$ on predictions $\bm{Y}$ follow two causal links: $ \bm{X} \rightarrow \bm{C} \rightarrow \bm{Y} $ and $ \bm{X} \rightarrow \bm{S} \rightarrow \bm{Y}$. These two causal links represent the CAER models' pure estimation of $\bm{Y}$ based on the total context representations $\bm{C}$ and subject representations $\bm{S}$ learned from $\bm{X}$. In practical implementations, $\bm{C}$ and $\bm{S}$ are generally integrated to serve the final emotion predictions in a joint manner, \eg, feature concatenation integration~\cite{mittal2020emoticon}.

According to the causal theory~\cite{pearl2009causal}, 
the confounder $\bm{Z}$ is the common cause of the input images $\bm{X}$ and
corresponding predictions $\bm{Y}$. 
The positive effects from context and subject features are reflected upon causal paths $ \bm{X} \rightarrow \bm{C} \rightarrow \bm{Y} $ and $ \bm{X} \rightarrow \bm{S} \rightarrow \bm{Y}$, respectively, which provide beneficial semantic information for recognition purposes.
Unfortunately, the confounder $\bm{Z}$  causes the negative effect of misleading the models to focus on spurious correlations instead of pure causal relationships.
This deleterious effect is propagated through a backdoor path $\bm{X} \leftarrow \bm{Z} \rightarrow \bm{C} \rightarrow \bm{Y}$ built with $\bm{Z}$ as the mediator, which we aim to prevent.

\subsection{Causal Intervention via Backdoor Adjustment}
We have now clarified the causal relationships among the CAER variables based on the aforementioned explanations. 
As shown in Figure~\ref{indro3}(a), the predictions of emotion probabilities from existing methods follow the likelihood estimation $P(\bm{Y}|\bm{X})$.
This process is formulated by the Bayes rule:
\begin{equation}
\begin{aligned}
P(\bm{Y}|\bm{X})=\sum_{\bm{z}}^{} P(\bm{Y}|\bm{X},\bm{S}=f_{s} (\bm{X}), \\
\bm{C}=f_{c} (\bm{X}, \bm{z})) P(\bm{z}|\bm{X}),
\label{one}
\end{aligned}
\end{equation}
where $f_{s} (\cdot)$ and $f_{c} (\cdot)$ are two generalized encoding functions that obtain the total $\bm{S}$ and $\bm{C}$, respectively. 
The backdoor confounder $\bm{z} \in \bm{Z}$ introduces the observational context bias through the conditional probability $ P(\bm{z}|\bm{X})$.
Theoretically, our goal is to remove confounding interference from $\bm{Z}$ and force the models to achieve unbiased predictions by relying only on valuable effects from $\bm{X}$ to $\bm{Y}$, \ie, $ \bm{X} \rightarrow \bm{C}/\bm{S} \rightarrow \bm{Y}$. The intuitive insight is implementing an intervention on $\bm{X}$ to enable the models to treat all context semantics fairly during training without favoring any of them. This philosophy can be viewed as conducting a
randomized controlled experiment by collecting images of subjects with any emotion in any context.
Nevertheless, this intervention is impractical since different subjects are situated in countless context scenarios in the real world, and enumerating them is difficult. To address this, we perform the causal intervention $P(\bm{Y}|do(\bm{X}))$ via the backdoor adjustment principle~\cite{pearl2009causal} to interrupt the unfavorable effect propagated along the backdoor path between $\bm{X}$ and $\bm{Y}$, where $do(\cdot)$ operator is an effective approximation for the imaginative intervention \cite{glymour2016causal}. According to the backdoor adjustment, we stratify $\bm{Z}$ to measure the causal effect. This operation means partitioning the contexts into homogeneous groups with respect to $\bm{Z}$ and then estimating the average causal effect by computing a weighted average based on the proportion of samples containing different context prototypes in the training data.
Thus, the models would approximate the causal effects through the intervention $P(\bm{Y}|do(\bm{X}))$ rather than capture spurious correlations through the likelihood $P(\bm{Y}|\bm{X})$. As shown in Figure~\ref{indro3}(b), the backdoor path would be invalid because the link from $\bm{Z}$ to $\bm{X}$ is cut off. 
After implementing the intervention in the new graph, Equation~(\ref{one}) is represented as follows by the Bayes rule:
\begin{equation}
\begin{aligned}
P(\bm{Y}|do(\bm{X}))=\sum_{\bm{z}}^{} P(\bm{Y}|\bm{X},\bm{S}=f_{s} (\bm{X}), \\
\bm{C}=f_{c} (\bm{X}, \bm{z})) P(\bm{z}).
\label{two}
\end{aligned}
\end{equation}
Since $\bm{z}$ is no longer influenced by $\bm{X}$, the intervention deliberately encourages $\bm{X}$ to fairly account for the effects of each $\bm{z}$ when predicting $\bm{Y}$. $P(\bm{z})$ is the priori probability that depicts the proportion of each $\bm{z}$ in the whole.

 \begin{figure*}[t]
  \centering
  \includegraphics[width=0.85\linewidth]{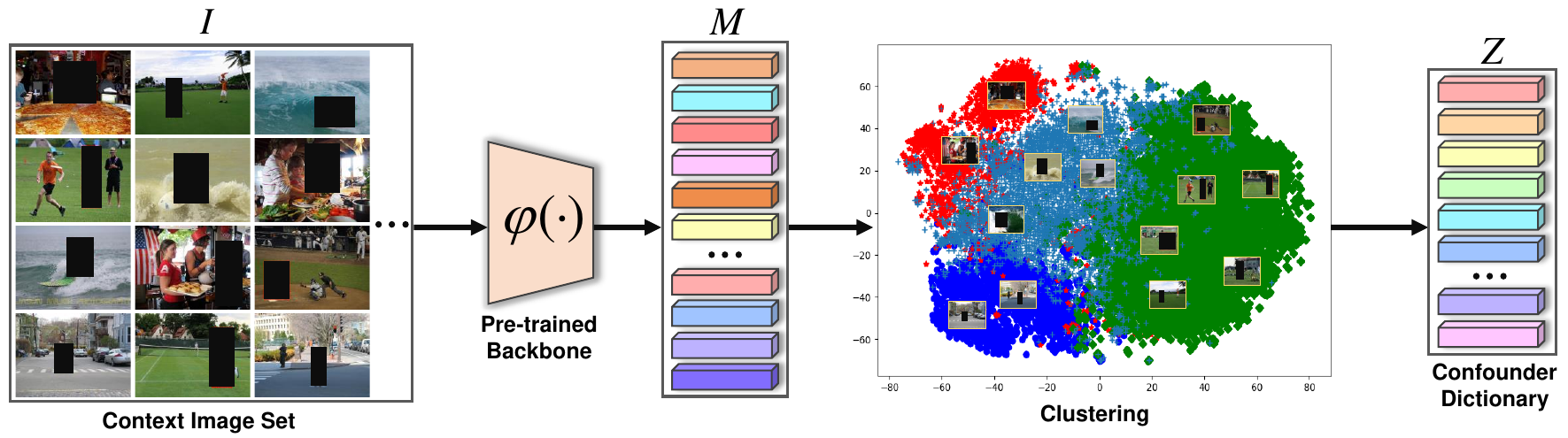}
  \caption{We show the generation procedure framework of the confounder dictionary $\bm{Z}$. The context image set $I$ is first generated by masking the recognized subject in each original training sample. Subsequently, the image set is fed to a pre-trained backbone $\varphi(\cdot) $ to extract the corresponding context representations and generate a context feature set $\bm{M}$. Ultimately, we utilize a clustering algorithm to learn different context prototypes and obtain the confounder dictionary $\bm{Z}$.
  }
  \label{clustering}
\end{figure*}

\subsection{Context-Deconfounded Training with  CCIM} 
To achieve the theoretical intervention in Equation~(\ref{two}) at the implementation level, we present a Contextual Causal Intervention Module (CCIM) for context-deconfounded training of CAER models.
From Figure~\ref{intervention}, CCIM is incorporated into the general pipeline of existing methods in a plug-and-play and model-agnostic manner. The output of CCIM is used in the task-specific classifier (\ie, neurons with the number of emotion categories) to perform the final predictions. The implementation of CCIM is described as follows.

\subsubsection{Confounder Dictionary}
Since collecting all contexts in the real world is impossible and there is a lack of supervised contextual information in the training data, we approximate a stratified confounder dictionary $\bm{Z}= [\bm{z}_1,\bm{z}_2, \dots,\bm{z}_{N} ]$ over the whole training samples using an unsupervised approach. $N$ is a size hyperparameter that represents the possible confounder number. Each $\bm{z}_i \in \mathbb{R}^{d}$ stands for a kind of context prototype in a stratified homogeneous group. As Figure~\ref{clustering} shows, we first mask the recognized subjects based on their priori bounding boxes to produce a context image set $I$. The masking operation aims to preserve only context-dependent regions to prevent subject-based attributes from impacting the construction of the confounder dictionary. We discuss its necessity in the experimental part. Concretely, for a given input image $\bm{x}$, its corresponding context image $I_x$ is expressed as follows:
\begin{equation}
    I_{x}= \begin{cases}x(i, j) & \text { if } x(i, j) \notin \text {bbox}_{\text {subject }}, \\ 0 & \text { otherwise }, \end{cases}
\label{bbx}
\end{equation}
where $\text{bbox}_\text{subject}$ means the bounding box of the recognized subject. Then, we employ a candidate pre-trained network $\varphi(\cdot)$ to generate the context feature set $\bm{M}=\{\bm{m}_{k} \in \mathbb{R}^{d}  \}^{N_{m}}_{k=1} $ from the context image set $I$,  where $N_m$ is the number of training samples. 
For flexibly obtaining context prototypes, we utilize unsupervised K-Means++ to learn $\bm{Z}$ so that each $\bm{z}_i$ represents a form of context cluster. Each $\bm{z}_i$ is set to the average feature from each cluster that aggregates the homogeneous confounding characteristics, which is expressed as follows:
\begin{equation}
\bm{z}_i = \frac{1}{N_i}  {\sum_{j=1}^{N_i}} \bm{m}^{i} _j,
\end{equation}
where $N_i$ is the number of context features in the $i$-th cluster. Note that there is no specific requirement for the choice of the clustering algorithm here, which we give justification in the subsequent ablation study.

\subsubsection{Parameterization of the Proposed CCIM}
According to Equation~(\ref{two}), the computation for $P(\bm{Y}|do(\bm{X}))$ is very expensive since we need to forward process each pair of $\bm{X}$ and $\bm{z}$ multiple times. To efficiently implement the causal intervention, we apply the Normalized Weighted Geometric Mean (NWGM)~\cite{xu2015show} to allow approximating the above expectation at the feature level:
\begin{equation}
\begin{aligned}
P(\bm{Y}|do(\bm{X})) \overset{\small{\text{NWGM}}}{\approx} P(\bm{Y}|\bm{X},\bm{S}=f_{s} (\bm{X}), \\
\bm{C}= \sum_{\bm{z}}^{}f_{c} (\bm{X}, \bm{z}) P(\bm{z})).
\label{three}
\end{aligned}
\end{equation}
Here, we instantiate a  parameterized network to efficiently approximate the conditional probability in Equation~(\ref{three}):
\begin{equation}
P(\bm{Y}|do(\bm{X})) = \bm{W}_{h}\bm{h}+\bm{W}_{g}\mathbb{E}_{\bm{z}}[g(\bm{z})],
\label{four}
\end{equation}
where $\bm{W}_{h}\in \mathbb{R}^{d_{m} \times d_{h}}$ and $\bm{W}_{g}\in \mathbb{R}^{d_{m} \times d}$ are the learnable parameters. $\bm{h}=\phi (\bm{s}, \bm{c})  \in \mathbb{R}^{d_{h} \times 1}$, and $\phi (\cdot)$ is a fusion strategy (\emph{e.g}., concatenation) that integrates $\bm{s}$ and $\bm{c}$ into the joint representation $\bm{h}$.
The above approximation implies that the output expectation for all possible confounders $\bm{z}$ can be calculated simply by feed-forward propagation with the expectation vector $\mathbb{E}_{\bm{z}}[g(\bm{z})]$ as the input.
Specifically, $\mathbb{E}_{\bm{z}}[g(\bm{z})]$ is approximated as a weighted integration of all context prototypes referencing the corresponding proportion:
\begin{equation}
\mathbb{E}_{\bm{z}}[g(\bm{z})]=
\sum_{i=1}^{N} \lambda_{i} \bm{z}_{i} P(\bm{z}_{i}),
\label{five}
\end{equation}
where $P(\bm{z}_{i})= \frac{N_i}{N_m}$ and $\lambda_{i}$ is a weighted attention score to measure the importance of the corresponding $\bm{z}_{i}$.
In practice, the integrated feature $\bm{h}$ from one sample queries each $\bm{z}_{i}$ in the confounder dictionary $\bm{Z} \in \mathbb{R}^{N \times d}$ to obtain the sample-specific attention set $\{\lambda_{i}  \}_{i=1}^{N}$. The intuitive insight is that each sample is impacted to varying degrees of distinct $\bm{z}_{i}$.
We provide two implementation patterns for $\lambda_{i}$. The first one is the dot product attention:
\begin{equation}
   \text{Dot Product}:  \lambda_{i} = softmax(\frac{(\bm{W}_{q} \bm{h})^{T} (\bm{W}_{k} \bm{z}_{i})}
    {\sqrt{d}} ),
    \label{six}
\end{equation}
and the second one is the additive attention:
\begin{equation}
 \text{Additive}:  \lambda_{i} = softmax(\bm{W}_{t}^{T}
    \cdot
    Tanh(\bm{W}_{q}\bm{h}+\bm{W}_{k} \bm{z}_{i})),
    \label{seven}
\end{equation}
where $\bm{W}_{t} \in \mathbb{R}^{d_{n} \times 1}$, $\bm{W}_{q}  \in \mathbb{R}^{d_{n} \times d_{h}}$, and $\bm{W}_{k}  \in \mathbb{R}^{d_{n} \times d}$ are learnable mapping weights.

\section{Datasets and Evaluation Metrics}
\label{dataset}
Our experiments are conducted on three standard CAER datasets, including EMOTIC~\cite{kosti2019context}, CAER-S~\cite{lee2019context}, and GroupWalk~\cite{mittal2020emoticon} datasets.

\noindent \textbf{EMOTIC} is the first large-scale CAER benchmark that contains 23,571 images of 34,320 annotated subjects. The majority of the images come from unconstrained environments to provide rich data resources on different subjects in diverse context scenarios. The bounding box coordinates of each recognized subject are provided in the annotation file to give the location information. EMOTIC supports two types of emotion descriptors: 26 discrete emotion categories for multi-label classification and 3 continuous emotion dimensions for regression. The discrete categories consist of ``\textit{Affection, Anger, Anticipation, Aversion, Confidence, Disapproval, Disconnection, Disquietment, Doubt/Confusion, Embarrassment, Engagement, Esteem, Excitement, Fatigue, Fear, Happiness, Pain, Peace, Pleasure, Sadness, Sensitivity, Suffering, Surprise, Sympathy,} and \textit{Yearning}''. The continuous dimensions are annotated following the mainstream VAD emotional state model~\cite{mehrabian1995framework}, which consists of ``\textit{Valence, Arousal,} and \textit{Dominance}''. The values of each dimension are constrained to range from 1 to 10 to express different emotion intensities.
We adopt the standard dataset partitioning for a fair comparison, \ie, 70\% data in the training set, 10\% data in the validation set, and 20\% data in the testing set.

\noindent \textbf{CAER-S} contains 70k static images captured from video clips. These images record different subjects in indoor and outdoor scenarios from 79 TV shows to include diverse contextual elements. CAER-S supports multi-class classification of emotion labels, and its annotated emotion categories include ``\textit{Anger, Disgust, Fear, Happy, Sad, Surprise,} and \textit{Neutral}''. The training, validation, and testing samples are randomly partitioned in a ratio of 7:1:2 during utilization.

\noindent \textbf{GroupWalk} consists of 45 manually collected videos from real-world environments. The annotated subjects have visible faces and gaits in all videos.
A characteristic of GroupWalk is containing extensive agent flows and interactions for understanding the subjects' affective effluence in the group effect. The annotations provide discrete emotion categories to support implementing multi-label classification over ``\textit{Angry, Happy, Neutral}, and \textit{Sad}''. The dataset partitioning is categorized as 85\% training set and 15\% testing set.

\noindent \textbf{Evaluation Metrics.}
We utilize the Average Precision (AP) to evaluate the discrete results on the EMOTIC and GroupWalk datasets. The Average Absolute Error (AAE) is employed to evaluate the testing results of the continuous dimensions on the EMOTIC. In addition, we follow~\cite{li2021human} to deeply evaluate the performance of multi-label classification on the EMOTIC using the label-based metrics (C-F1) and sample-based metrics (O-F1). For the CAER-S, the standard classification accuracy is used for evaluation.

\begin{table}[t]
\setlength{\tabcolsep}{24pt}
\centering
\caption{Quantitative results of different methods and CCIM-based models on the EMOTIC dataset. We report the mean average precision (mAP) to provide comprehensive comparison experiments.  $\ast$ represents the results from the original reports. $\dagger$ represents the results from our implementation. $\uparrow$ represents the improvement of the CCIM-based version over the vanilla model. The improved results are marked in \textbf{bold}. The footnotes $\ast$, $\dagger$, and $\uparrow$ of Tables~\ref{emotic}, \ref{emotic_f1}, \ref{emotic_aae}, \ref{caer} and~\ref{groupwalk} follow the same interpretation.}
\resizebox{\linewidth}{!}{%
\begin{tabular}{c|c}
\toprule
\textbf{Methods}       & \textbf{mAP}~(\%)   \\ \midrule
HLCR~\cite{de2023high}          & $30.02^{\ast}$ \\
TEKG~\cite{chen2023incorporating}         & $31.36^{\ast}$ \\
RRLA~\cite{li2021human}          & $32.41^{\ast}$ \\
VRD~\cite{hoang2021context}           & $35.16^{\ast}$ \\
SIB-Net~\cite{li2021sequential}       & $35.41^{\ast}$ \\
MCA~\cite{yang2022emotion}           & $37.73^{\ast}$ \\ \midrule
EMOT-Net~\cite{kosti2019context}      & $27.93^{\dagger}$ \\
EMOT-Net +\textbf{ CCIM} & $\textbf{30.88}^{\dagger}$ (${ \uparrow2.95 }$ ) \\
CAER-Net~\cite{lee2019context}      & $23.85^{\dagger}$ \\
CAER-Net + \textbf{CCIM} & $\textbf{26.51}^{\dagger}$ (${ \uparrow2.66 }$ ) \\
GNN-CNN~\cite{zhang2019context}       & $28.16^{\dagger}$ \\
GNN-CNN + \textbf{CCIM}  & $\textbf{31.72}^{\dagger}$ (${ \uparrow3.56 }$ ) \\
CD-Net~\cite{wang2022context}        & $28.87^{\dagger}$ \\
CD-Net + \textbf{CCIM}   & $\textbf{32.29}^{\dagger}$ (${ \uparrow3.42 }$ ) \\
EmotiCon~\cite{mittal2020emoticon}      & $35.28^{\dagger}$ \\
EmotiCon + \textbf{CCIM} & $\textbf{39.13}^{\dagger}$  (${ \uparrow3.85 }$ ) \\ \bottomrule
\end{tabular}
}
\label{emotic_map}
\end{table}

\begin{table*}[t]
\setlength{\tabcolsep}{2pt}
\renewcommand{\arraystretch}{1.2}
\centering
\caption{ Quantitative results of CCIM-based models for each emotion category on the EMOTIC dataset. We report the average precision of each category to provide comprehensive comparison experiments. The improved results are marked in \textbf{bold}.}
\resizebox{\linewidth}{!}{%
\begin{tabular}{llcccccccccc}
\toprule
\multicolumn{2}{l}{\multirow{2}{*}{\textbf{Category}}} & \multirow{2}{*}{EMOT-Net} & \multirow{2}{*}{\begin{tabular}[c]{@{}c@{}}EMOT-Net\\ +~\textbf{CCIM}\end{tabular}} & \multirow{2}{*}{GNN-CNN} & \multirow{2}{*}{\begin{tabular}[c]{@{}c@{}}GNN-CNN\\ +~\textbf{CCIM}\end{tabular}} & \multirow{2}{*}{CAER-Net} & \multirow{2}{*}{\begin{tabular}[c]{@{}c@{}}CAER-Net\\ +~\textbf{CCIM}\end{tabular}} & \multirow{2}{*}{CD-Net} & \multirow{2}{*}{\begin{tabular}[c]{@{}c@{}}CD-Net\\ +~\textbf{CCIM}\end{tabular}} &  \multirow{2}{*}{EmotiCon} & \multirow{2}{*}{\begin{tabular}[c]{@{}c@{}}EmotiCon\\ +~\textbf{CCIM}\end{tabular}} \\
\multicolumn{2}{l}{}                          &                             &                                                                             &                          &                                                                          &                           &                                                                           &                          &                         &                           &                                                                           \\ \midrule
\multicolumn{2}{l}{Affection}   & 26.47       & 34.87    & 47.52     & 36.18  & 22.36    & 23.08                                          & 28.44                    & 30.25                   & 38.55                     & 40.77                                                                     \\
\multicolumn{2}{l}{Anger}                     & 11.24                       & 13.05                                                                        & 11.27                    & 12.53                                                                    & 12.88                     & 12.99                                                                     & 12.12                    & 14.31                   & 14.69                     & 15.48                                                                     \\
\multicolumn{2}{l}{Annoyance}                 & 15.26                       & 18.04                                                                       & 12.33                    & 13.73                                                                    & 14.42                     & 15.28                                                                     & 19.71                   & 22.76                   & 24.68                     & 24.47                                                                     \\
\multicolumn{2}{l}{Anticipation}              & 57.31                       & 94.19                                                             & 63.20                     & 92.32                                                           & 52.85                     & 90.03                                                            & 57.65                    & 91.80                   & 60.73                     & 95.15                                                            \\
\multicolumn{2}{l}{Aversion}                  & 7.44                        & 13.41                                                              & 6.81                     & 15.41                                                          & 3.26                      & 12.96                                                            & 9.94                     & 13.89                   & 11.33                     & 19.38                                                            \\
\multicolumn{2}{l}{Confidence}                & 80.33                       & 74.90                                                                        & 74.83                    & 75.01                                                                    & 72.68                     & 73.24                                                                     & 69.26                    & 68.17                   & 68.12                     & 75.81                                                                     \\
\multicolumn{2}{l}{Disapproval}               & 16.14                       & 19.87                                                                       & 12.64                    & 14.45                                                                    & 15.37                     & 16.38                                                                     & 22.78                    & 23.65                   & 18.55                     & 23.65                                                                     \\
\multicolumn{2}{l}{Disconnection}             & 20.64                       & 27.72                                                                       & 23.17                    & 30.52                                                                    & 22.01                     & 23.39                                                                     & 27.55                    & 31.52                   & 28.73                     & 31.93                                                                     \\
\multicolumn{2}{l}{Disquietment}              & 19.57                       & 19.12                                                                       & 17.66                    & 20.85                                                                    & 10.84                     & 18.10                                                                      & 21.04                    & 25.87                   & 22.14                     & 26.84                                                                     \\
\multicolumn{2}{l}{Doubt/Confusion}           & 31.88                       & 19.35                                                                       & 19.67                    & 20.43                                                                    & 26.07                     & 17.66                                                                     & 24.23                     & 19.46                   & 38.43                     & 34.28                                                                     \\
\multicolumn{2}{l}{Embarrassment}             & 3.05                        & 6.23                                                               & 1.58                     & 9.21                                                            & 1.88                      & 5.86                                                             & 4.50                     & 10.03                    & 10.31                     & 16.73                                                           \\
\multicolumn{2}{l}{Engagement}                & 86.69                       & 88.93                                                                       & 87.31                    & 96.88                                                                    & 73.71                     & 70.04                                                                     & 85.32                    & 82.55                   & 86.23                     & 97.41                                                                     \\
\multicolumn{2}{l}{Esteem}                    & 17.86                       & 21.69                                                              & 12.05                    & 22.72                                                           & 15.38                     & 16.67                                                            & 18.66                     & 22.00                   & 25.75                     & 27.44                                                            \\
\multicolumn{2}{l}{Excitement}                & 78.05                       & 73.81                                                                       & 72.68                    & 73.21                                                                    & 70.42                     & 71.08                                                                     & 70.07                    & 68.32                   & 80.75                     & 81.59                                                                     \\
\multicolumn{2}{l}{Fatigue}                   & 8.87                        & 9.96                                                                        & 12.93                    & 12.66                                                                    & 6.29                      & 9.73                                                                      & 11.56                    & 14.97                   & 19.35                     & 15.53                                                                     \\
\multicolumn{2}{l}{Fear}                      & 15.70                        & 9.04                                                                        & 6.15                     & 10.31                                                                    & 7.47                      & 6.61                                                                      & 10.38                    & 6.85                   & 16.99                     & 15.37                                                                     \\
\multicolumn{2}{l}{Happiness}                 & 58.92                       & 78.09                                                             & 72.90                     & 75.64                                                           & 53.73                     & 62.34                                                            & 68.46                    & 76.11                   & 80.45                     & 83.55                                                            \\
\multicolumn{2}{l}{Pain}                      & 9.46                        & 14.71                                                              & 8.22                     & 15.36                                                           & 8.16                      & 9.43                                                             & 13.82                    & 17.26                   & 14.68                     & 17.76                                                            \\
\multicolumn{2}{l}{Peace}                     & 22.35                       & 22.79                                                                       & 30.68                    & 23.88                                                                    & 19.55                     & 20.21                                                                     & 28.18                   & 25.08                   & 35.72                     & 38.94                                                                     \\
\multicolumn{2}{l}{Pleasure}                  & 46.72                       & 46.59                                                                       & 48.37                    & 45.52                                                                    & 34.12                     & 35.37                                                                     & 47.64                    & 50.60                   & 67.31                     & 64.57                                                                     \\
\multicolumn{2}{l}{Sadness}                   & 18.69                       & 17.47                                                                       & 23.90                     & 22.08                                                                    & 17.75                     & 13.24                                                                     & 32.99                     & 36.24                   & 40.26                     & 45.63                                                                     \\
\multicolumn{2}{l}{Sensitivity}               & 9.05                        & 7.91                                                                        & 4.74                     & 8.02                                                                     & 6.94                      & 4.74                                                                      & 7.21                     & 7.08                   & 13.94                     & 17.04                                                                     \\
\multicolumn{2}{l}{Suffering}                 & 17.67                       & 15.35                                                                       & 23.71                    & 18.45                                                                    & 14.85                     & 11.89                                                                     & 35.19                     & 28.99                   & 48.05                     & 21.52                                                                     \\
\multicolumn{2}{l}{Surprise}                  & 22.38                       & 13.12                                                                       & 8.44                     & 13.93                                                                    & 17.46                     & 11.70                                                                      & 7.42                    & 10.25                   & 19.60                      & 26.81                                                                     \\
\multicolumn{2}{l}{Sympathy}                  & 15.23                       & 32.60                                                              & 19.45                    & 33.95                                                           & 14.89                     & 28.59                                                            & 10.33                    & 32.46                   & 16.74                     & 47.60                                                             \\
\multicolumn{2}{l}{Yearning}                  & 9.22                        & 10.08                                                                        & 9.86                     & 11.58                                                                    & 4.84                      & 8.61                                                                     & 6.24                    & 9.17                   & 15.08                     & 12.25                                                                     \\ \midrule
\multicolumn{2}{l}{\textbf{mAP} (\%)}              & $27.93^{\dagger}$                       & $\textbf{30.88}^{\dagger}$                                                            & $28.16^{\dagger}$                    & $\textbf{31.72}^{\dagger}$                                                             & $23.85^{\dagger}$                     & $\textbf{26.51}^{\dagger}$                                                            & $28.87^{\dagger}$                    & $\textbf{32.29}^{\dagger}$                    & $35.28^{\dagger}$                     & $\textbf{39.13}^{\dagger}$                                                                      \\ \bottomrule
\end{tabular}
}
\label{emotic}
\end{table*}

\section{Implementation Details}
\label{detail}
\subsection{Model Zoo}
While current studies offer promising advancements, most efforts are not open-source, and the design philosophies in several approaches are similar. In this situation, we choose five representative approaches that include classical and state-of-the-art (SOTA) works. The selected approaches have entirely different network structures and modeling paradigms to support an exhaustive evaluation of the effectiveness and applicability of the proposed CCIM.
A brief introduction to these approaches is given below.

\noindent \textbf{EMOT-Net}~\cite{kosti2019context} is a classical Convolutional Neural Network (CNN) model. The model has two different branches, one for extracting physical features from the recognized subject region and the other for extracting contextual semantics from the global background region.

\noindent \textbf{GNN-CNN}~\cite{zhang2019context} extracts subject body information using standard CNN network.
Moreover, the context-related elements extracted via the region proposal network~\cite{ren2015faster} are considered as nodes and infer the emotional states via the Graph Neural Network (GNN).

\noindent \textbf{CAER-Net}~\cite{lee2019context} consists of two CNN encoding networks and an adaptive fusion module. The two encoders extract information from the subject's face and emotional cues from the background context after masking the face.

\noindent \textbf{CD-Net}~\cite{wang2022context} first obtains intermediate features for face, body, and context regions via ResNet~\cite{he2016deep}. Then, a tubal transformer is designed to facilitate fine-grained interactions and hierarchical fusion across multi-scale features.

\noindent \textbf{EmotiCon}~\cite{mittal2020emoticon} is a multi-stream model with three context-dependent branches. The subject-centered branch uses facial and gait keypoints to learn human dynamics. The out-of-subject context branches utilize visual attention and depth maps to capture specific emotion semantics.

We re-train EMOT-Net according to the official codebase. Meanwhile, we reproduce the results of other SOTA models (\ie, GNN-CNN, CAER-Net, CD-Net, and EmotiCon) based on the details provided in the original reports.

\subsection{Confounder Construction}
One of the vital steps in the confounder construction is to locate the recognized subject and remove the influence from the subject's information. To this end, we use the pre-trained Faster R-CNN~\cite{ren2015faster} to detect the bounding boxes of the recognized subject for each training sample on both CAER-S and GroupWalk. EMOTIC provides annotated information about the bounding box for utilization. We then utilize the bounding boxes to mask the target subjects according to Equation~(\ref{bbx}) for producing the context image set $I$. After that, we employ the ResNet-152~\cite{he2016deep} pre-trained on the 
Places365~\cite{zhou2017places} dataset to extract the context feature set $\bm{M}$.
The rich scene context resources in the Places365 dataset facilitate distilling informative context semantics from the pre-training backbone and learning better context prototypes in the subsequent clustering process.
The final pooling layer is applied to obtain each context feature $\bm{m}$ for retaining the refined feature semantics.
The hidden feature dimension $d$ is set to 2048. 
The default size $N$ (\ie, the number of clusters) is set to 256, 128, and 256 in the EMOTIC, CAER-S, and GroupWalk datasets, respectively.

\subsection{Training Details}
All reproduced models and our CCIM are implemented via the PyTorch toolbox~\cite{paszke2017automatic}. The computational resources utilize Nvidia Tesla V100 GPUs. Note that CCIM as a plug-and-play component does not affect the training protocols of the original methods. For this reason, we adopt precisely the identical training details provided by the original models to ensure a fair comparison.
To implement our CCIM, the hidden dimensions $d_{m}$ and $d_{n}$ are set to 128 and 256, respectively.
The output dimension $d_{h}$ of the joint feature $\bm{h}$ in the different approaches is 256 (EMOT-Net), 1024 (GNN-CNN), 128 (CAER-Net), 512 (CD-Net), and 78 (EmotiCon).

\subsection{Comparison with State-of-the-art Methods}
To comprehensively evaluate the performance of the proposed CCIM, we compare the CCIM-based models with existing SOTA methods, including HLCR~\cite{de2023high}, TEKG~\cite{chen2023incorporating}, RRLA~\cite{li2021human}, VRD~\cite{hoang2021context}, SIB-Net~\cite{li2021sequential}, MCA~\cite{yang2022emotion}, CAGBN~\cite{ruan2020context}, and GRERN \cite{gao2021graph}. 
The dot product attention of Equation~(\ref{six}) is used as the default implementation.
on of \ref{six}.

\subsubsection{Quantitative Results on the EMOTIC}
As a holistic benchmark in the CAER field, the EMOTIC dataset features assessment patterns on 26 discrete categories and 3 continuous dimensions (VAD) of emotions. To this end, we provide systematic experiments in both directions. For discrete categories, we have the following core observations. \textbf{(i)} Table~\ref{emotic_map} first presents the mean average precision (mAP) results across all emotion categories to support the macroscopic evaluation. CCIM significantly brings consistent performance gains to existing models and achieves new SOTAs.
Concretely, the CCIM-based EMOT-Net, CAER-Net, GNN-CNN, CD-Net, and EmotiCon improve the mAP scores by 2.95\%, 2.66\%, 3.56\%, 3.42\%, and 3.85\%, respectively, outperforming the vanilla methods by large margins.
\textbf{(ii)} HLCR and TEKG attempt to incorporate external knowledge and enhance emotion perception through linguistic semantic descriptors extracted from images.
Despite promising solutions, they suffer from severe performance bottlenecks since linguistic information produced from confounded contexts is essentially an implicit amplification of the adverse effects of the confounder. In comparison, the baselines (\eg, EMOT-Net, GNN-CNN, and CD-Net) improved by CCIM exhibit competitive or better results. Based on this finding, the key to solving the vision-driven CAER task is disentangling the underlying causal dependencies rather than fancy resorting to language community development.
\textbf{(iii)} Compared to current SOTA works (\eg, SIB-Net and MCA) with complex module stacking and massive parameters, EmotiCon achieves the best performance with the mAP score of 39.13\% only through the lightweight CCIM. This observation further demonstrates the effectiveness of our component.

Microscopically, we show the average precision (AP) scores for the CCIM-based models and their vanilla counterparts on each emotion category in Table~\ref{emotic} to provide more in-depth analyses.
\textbf{(i)} CCIM consistently improves performance in all methods for most emotion categories. For instance,  CCIM yields an average gain of 8.25\% on the AP scores across the five models for the ``\textit{Happiness}'' category reflecting positivity. Meanwhile, CCIM provides an average gain of 4.04\% on the AP scores across the five models for the ``\textit{Pain}'' category reflecting negativity.
These results imply that our component can effectively mitigate the performance bottleneck caused by the uneven distribution of emotion semantics of different polarities in context-based visual scenarios.
\textbf{(ii)} Moreover, CCIM remarkably improves the AP scores of some categories heavily persecuted by the confounder. For example, These CCIM-based methods boosted the AP scores by 29\%$\sim$37\% and 14\%$\sim$29\% on the ``\textit{Anticipation}'' and ``\textit{Sympathy}'' categories, respectively, significantly superior to their original models.
\textbf{(iii)} Due to adverse effects from the context bias, the performance of most models is usually poor on infrequent categories, such as ``\textit{Aversion}'' (AP scores of about 3\%$\sim$11\%) and ``\textit{Embarrassment}'' (AP scores of about 1\%$\sim$10\%). Thanks to the proposed CCIM, the AP scores in these two categories are achieved at about 12\%$\sim$19\% and 5\%$\sim$16\%.

\begin{table}[t]
\setlength{\tabcolsep}{10pt}
\centering
\caption{Quantitative results of different methods and CCIM-based models on the EMOTIC dataset. We report the Label-based F1 (C-F1) and example-based F1 (O-F1) scores to provide comprehensive comparison experiments. The improved results are marked in \textbf{bold}.}
\resizebox{\linewidth}{!}{%
\begin{tabular}{c|cc}
\toprule
\textbf{Methods}        & \textbf{C-F1}~(\%)  & \textbf{O-F1}~(\%)  \\ \midrule
ML-KNN~\cite{zhang2007ml}         & $6.57^{\ast}$  & 26.67$^{\ast}$ \\
Label Powerset~\cite{tsoumakas2010random} & 7.66$^{\ast}$  & 37.20$^{\ast}$  \\ \midrule
CAGBN~\cite{ruan2020context}          & 13.42$^{\ast}$ & 45.77$^{\ast}$ \\
RRLA~\cite{li2021human}           & 15.10$^{\ast}$  & 48.07$^{\ast}$ \\ 
EMOT-Net~\cite{kosti2019context}       & 8.27$^{\dagger}$  & 39.84$^{\dagger}$ \\
EMOT-Net + \textbf{CCIM}  & \textbf{9.35}$^{\dagger}$ (${\uparrow1.08}$)  & \textbf{43.38}$^{\dagger}$ (${\uparrow3.54}$) \\
CAER-Net~\cite{lee2019context}       & 7.14$^{\dagger}$  & 34.03$^{\dagger}$ \\
CAER-Net + \textbf{CCIM}  & \textbf{7.22}$^{\dagger}$ (${\uparrow0.08}$)  & \textbf{36.41}$^{\dagger}$ (${\uparrow2.38}$) \\
GNN-CNN~\cite{zhang2019context}        & 12.47$^{\dagger}$ & 42.55$^{\dagger}$ \\
GNN-CNN + \textbf{CCIM}   & \textbf{12.86}$^{\dagger}$ (${\uparrow0.39}$) & \textbf{46.34}$^{\dagger}$ (${\uparrow3.79}$) \\
CD-Net~\cite{wang2022context}         & 13.29$^{\dagger}$ & 41.95$^{\dagger}$ \\
CD-Net + \textbf{CCIM}    & 11.66$^{\dagger}$ & \textbf{44.72}$^{\dagger}$ (${\uparrow2.77}$) \\
EmotiCon~\cite{mittal2020emoticon}       & 13.59$^{\dagger}$ & 46.64$^{\dagger}$ \\
EmotiCon + \textbf{CCIM}  & \textbf{15.01}$^{\dagger}$(${\uparrow1.42}$) & \textbf{48.18}$^{\dagger}$(${\uparrow2.14}$) \\ \bottomrule
\end{tabular}
}
\label{emotic_f1}
\end{table}

Evaluating the EMOTIC dataset from the multi-label learning (MLL) perspective is an emerging paradigm due to the intrinsic connections among multiple emotion labels. Following the validation metrics~\cite{ruan2020context} of MLL, we adopt the Label-based F1 (C-F1) and example-based F1 (O-F1) scores in Table~\ref{emotic_f1} to measure the performance of the accessible methods and the CCIM-based models, where the average is taken over all classes and all testing examples, respectively. Some key findings are as follows.
\textbf{(i)} Deep learning-driven works usually obtain better results than machine learning-based efforts (\ie, ML-KNN and Label Powerset), suggesting that traditional efforts fail to capture profound dependencies across emotion categories.
\textbf{(ii)} CCIM yields considerable gains for most reproduced models. For instance, the O-F1 scores increased by an average of 2.92\% across the five models.
\textbf{(iii)} Despite the competitive results achieved by CAGBN and RRLA under the MLL evaluation scheme through the multi-label dependency modeling, they either rely on an incremental sequence generation~\cite{ruan2020context} or require additional topological guidance~\cite{li2021human}, causing sub-optimal solutions. In comparison, the CCIM-based EmotiCon achieves comparable or better performance with C-F1 and O-F1 scores of 15.01\% and 48.18\%, demonstrating the superiority of our component.

\begin{table}[t]
\setlength{\tabcolsep}{6pt}
\renewcommand{\arraystretch}{1.2}
\centering
\caption{Continuous dimension results of CCIM-based models on the EMOTIC dataset. We report the Average Absolute Error (AAE) scores to provide comparison experiments. The improved results are marked in \textbf{bold}.}
\resizebox{\linewidth}{!}{%
\begin{tabular}{c|ccc|c}
\toprule
\textbf{Methods}       & \textbf{Valence} & \textbf{Arousal} & \textbf{Dominance} & \textbf{Mean}   \\ \midrule
EMOT-Net~\cite{kosti2019context}      & 0.0533  & 0.0605  & 0.0576    & 0.0571 \\
EMOT-Net + \textbf{CCIM} & \textbf{0.0530}   & \textbf{0.0586}  & \textbf{0.0561}    & \textbf{0.0559} \\
GNN-CNN~\cite{zhang2019context}       & 0.0516  & 0.0571  & 0.0578    & 0.0555 \\
GNN-CNN + \textbf{CCIM}  & \textbf{0.0508}  & \textbf{0.0563}  & \textbf{0.0542}    & \textbf{0.0538} \\ \bottomrule
\end{tabular}
}
\label{emotic_aae}
\end{table}

Table~\ref{emotic_aae} reports quantitative results on the continuous dimensions of VAD for emotional states using the AAE scores (the lower, the better). For the VAD model, \textit{Valence} measures the degree of negativity or positivity of a subject's emotion. \textit{Arousal} measures a subject's agitation level, usually from inactive to ready for action. \textit{Dominance} measures a subject's control level over the situation, usually from uncontrolled to totally dominant. We only evaluate EMOT-Net and GNN-CNN due to other implemented models that do not support the regression task.
\textbf{(i)} Overall, CCIM consistently reduces the prediction errors on the three emotion dimensions due to producing lower AAE results, implying the applicability and effectiveness of our component over different emotional state spaces.
\textbf{(ii)} An interesting phenomenon is that improvements on \textit{Arousal} and \textit{Dominance} are significantly better than those on \textit{Valence}. A plausible explanation is that subjects generally show pronounced differences in agitation and control levels across distinct contexts, which are more vulnerable to the poison of context bias. For instance, inactive subjects with lower numerical values of \textit{Arousal} are usually located in similar indoor scenarios. Conversely, subjects with higher \textit{Arousal} values are usually located outdoors in diverse venues. In this case, spurious correlations between similar contexts and specific emotion polarity are more likely to be established, causing harmful performance bottlenecks. Fortunately, CCIM reasonably mitigates the detrimental effects and helps existing models achieve better overall performance (mean scores of 0.0559 on EMOT-Net and 0.0538 on GNN-CNN).

\begin{table}[t]
\setlength{\tabcolsep}{24pt}
\centering
\caption{ Quantitative results of different methods and CCIM-based models on the CAER-S dataset. We report the classification accuracy to provide comparison experiments. The improved results are marked in \textbf{bold}.}
\resizebox{\linewidth}{!}{%
\begin{tabular}{c|c}
\toprule
\textbf{Methods}            & \textbf{Accuracy}~(\%) \\ \midrule
Fine-tuned AlexNet~\cite{krizhevsky2012imagenet} & $61.73^{\ast}$    \\
Fine-tuned VGGNet~\cite{simonyan2014very}  & $64.85^{\ast}$    \\
Fine-tuned ResNet~\cite{he2016deep}  & $68.46^{\ast}$    \\
SIB-Net~\cite{li2021sequential}            & $74.56^{\ast}$    \\
MCA~\cite{yang2022emotion}                & $79.57^{\ast}$    \\
GRERN~\cite{gao2021graph}              & $81.31^{\ast}$    \\
RRLA~\cite{li2021human}               & $84.82^{\ast}$    \\
VRD~~\cite{hoang2021context}                & $90.49^{\ast}$    \\ \midrule
EMOT-Net~\cite{kosti2019context}           & $74.51^{\dagger}$    \\
EMOT-Net + \textbf{CCIM}      & $\textbf{75.82}^{\dagger}$ (${ \uparrow1.31 }$)     \\
CAER-Net~\cite{lee2019context}           & $73.47^{\dagger}$    \\
CAER-Net + \textbf{CCIM}      & $\textbf{74.81}^{\dagger}$ (${ \uparrow1.34 }$)     \\
GNN-CNN~\cite{zhang2019context}            & $77.21^{\dagger}$    \\
GNN-CNN + \textbf{CCIM}       & $\textbf{78.66}^{\dagger}$ (${ \uparrow1.45 }$)     \\
CD-Net~\cite{wang2022context}             & $85.33^{\dagger}$    \\
CD-Net + \textbf{CCIM}        & $\textbf{86.61}^{\dagger}$ (${ \uparrow1.28 }$)     \\
EmotiCon~\cite{mittal2020emoticon}           & $88.65^{\dagger}$    \\
EmotiCon + \textbf{CCIM}      & $\textbf{91.17}^{\dagger}$ (${ \uparrow2.52}$)   \\ \bottomrule
\end{tabular}
}
\label{caer}
\end{table}

\subsubsection{Quantitative  Results on the CAER-S}
\begin{figure*}[t]
  \centering
  \includegraphics[width=\linewidth]{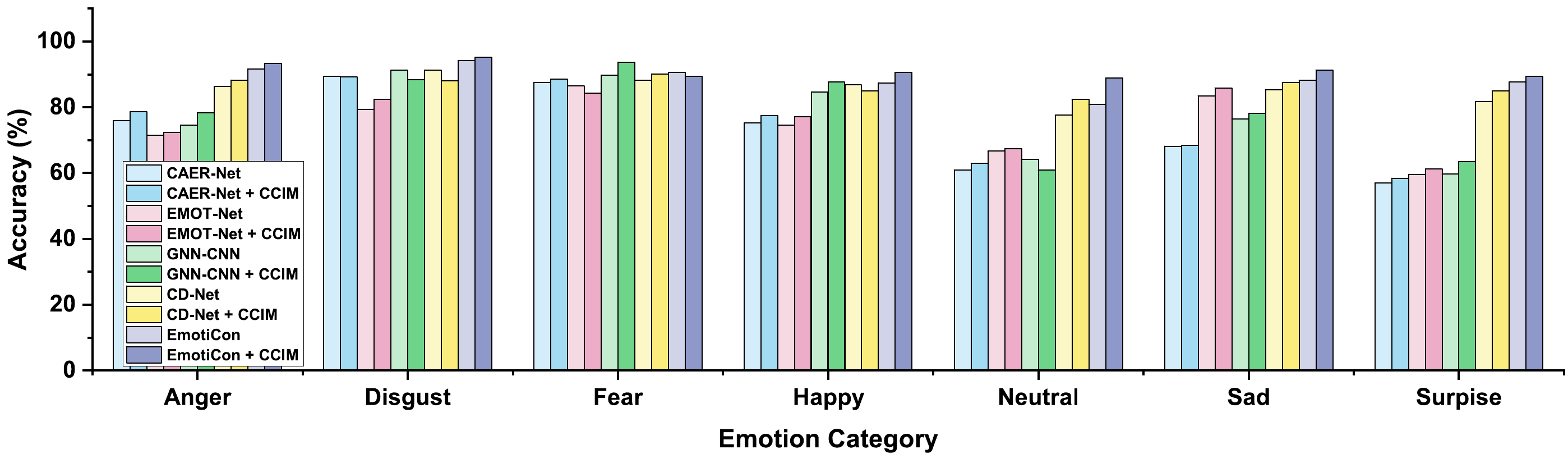}
  \caption{Emotion classification accuracy (\%) for each category of different CCIM-based models on the CAER-S dataset.
  }
  \label{accuracy}
\end{figure*}

Table~\ref{caer} provides the overall accuracy of different methods and CCIM-based models on the CAER-S dataset. Some key observations and analyses are as follows.
\textbf{(i)} Fine-tuned conventional models (\ie, AlexNet, VGGNet, and ResNet) typically have restricted performance upper bounds since the results usually do not exceed 70\%, implying a failure to capture adequate emotion semantics.
\textbf{(ii)} The proposed CCIM comprehensively enhances the performance of EMOT-Net, CAER-Net, GNN-CNN, and CD-Net. Compared to the vanilla models, their CCIM-based versions are improved by 1.31\%, 1.34\%, 1.45\%, and 1.28\%, respectively. The potential deduction is that CCIM forces each context prototype extracted from TV show scenarios to reasonably incorporate into the emotion predictions and improve the overall accuracy.
\textbf{(iii)} More importantly, the CCIM-based EmotiCon obtains the most significant gain of 2.52\% while beating all existing methods with an accuracy of 91.17\%.
\textbf{(iv)} Furthermore, we show the classification accuracy of each emotion category from different CCIM-based models on the CAER-S dataset in Figure~\ref{accuracy}. Overall, all models obtain considerable performance gains in most categories.

\subsubsection{Quantitative  Results on the GroupWalk}
As shown in Table~\ref{groupwalk}, our CCIM effectively improves the performance of EMOT-Net, CAER-Net, GNN-CNN, CD-Net, and EmotiCon on the GroupWalk dataset for most categories. The mAP scores for these models are increased by 2.41\%, 2.25\%, 2.99\%, 2.72\%, and 3.73\%, respectively. A noteworthy observation is that the \textit{``Neutral''} category exhibits slight deterioration across different models. The potential reason may be that samples with neutral emotions are more dispersed across contexts than samples with other emotions, leading to insufficient confounding effects.
Consequently, our component may experience slight over-intervention when decoupling the spurious ``\textit{context-emotion}'' mapping.
However, the minor sacrifice is tolerable compared to the overall superiority of the proposed CCIM.

\begin{table*}[t]
\setlength{\tabcolsep}{0.5pt}
\renewcommand{\arraystretch}{1.2}
\centering
\caption{Quantitative results of CCIM-based models for each emotion category on the GroupWalk dataset. We report the average precision of each category to provide comprehensive comparison experiments. The improved results are marked in \textbf{bold}.}
\resizebox{\linewidth}{!}{%
\begin{tabular}{c|cccccccccc}
\toprule
\multirow{2}{*}{\textbf{Category}} & \multirow{2}{*}{EMOT-Net} & \multirow{2}{*}{\begin{tabular}[c]{@{}c@{}}EMOT-Net\\  + \textbf{CCIM}\end{tabular}} & \multirow{2}{*}{CAER-Net} & \multirow{2}{*}{\begin{tabular}[c]{@{}c@{}}CAER-Net \\ + \textbf{CCIM}\end{tabular}} & \multirow{2}{*}{GNN-CNN} & \multirow{2}{*}{\begin{tabular}[c]{@{}c@{}}GNN-CNN \\ + \textbf{CCIM}\end{tabular}} & \multirow{2}{*}{CD-Net} & \multirow{2}{*}{\begin{tabular}[c]{@{}c@{}}CD-Net \\ + \textbf{CCIM}\end{tabular}} & \multirow{2}{*}{EmotiCon} & \multirow{2}{*}{\begin{tabular}[c]{@{}c@{}}EmotiCon \\ + \textbf{CCIM}\end{tabular}} \\
                          &                           &                                                                             &                           &                                                                             &                          &                                                                            &                         &                                                                           &                           &                                                                             \\ \midrule
Angry                     & 57.65                     & 62.41                                                                       & 45.18                     & 50.43                                                                       & 51.92                    & 54.07                                                                      & 59.28                   & 64.35                                                                     & 68.85                     & 75.93                                                                       \\
Happy                     & 71.32                     & 75.68                                                                       & 56.59                     & 60.71                                                                       & 63.37                    & 70.25                                                                      & 74.06                   & 77.84                                                                     & 72.31                     & 79.15                                                                       \\
Neutral                   & 43.10                      & 41.03                                                                       & 39.32                     & 37.84                                                                       & 40.26                    & 39.49                                                                      & 45.43                   & 42.41                                                                     & 50.34                     & 48.66                                                                       \\
Sad                       & 61.24                     & 63.84                                                                       & 52.96                     & 54.06                                                                       & 58.15                    & 61.85                                                                      & 61.65                   & 66.72                                                                     & 70.80                      & 73.48                                                                       \\ \midrule
mAP                       & 58.33$^{\dagger}$                      & \textbf{60.74}$^{\dagger}$ (${\uparrow2.41}$)                                                               & 48.51$^{\dagger}$                      & \textbf{50.76}$^{\dagger}$ (${\uparrow2.25}$)                                                              & 53.43$^{\dagger}$                     & \textbf{56.42}$^{\dagger}$ (${\uparrow2.99}$)                                                              & 60.11$^{\dagger}$                    & \textbf{62.83}$^{\dagger}$ (${\uparrow2.72}$)                                                             & 65.58$^{\dagger}$                      & \textbf{69.31}$^{\dagger}$  (${\uparrow3.73}$)                                                              \\ \bottomrule
\end{tabular}
}
\label{groupwalk}
\vspace{-3pt}
\end{table*}

\subsubsection{Discussion from the Causal Perspective}
\label{dcp}

Besides the above observations and analysis, we have two critical insights from the causal perspective across all datasets.
\textbf{(i)} The performance improvements of all methods on the EMOTIC and GroupWalk datasets are more significant than on the CAER-S dataset. Taking the mAP and accuracy metrics as examples, the average gains across models on EMOTIC, CAER-S, and GroupWalk are 3.29\%, 1.59\%, and 2.82\%, respectively. 
Combined with Figure~\ref{indro2}, we realize that EMOTIC suffers a more severe context bias than the CAER-S dataset. These findings exhibit an encouraging conclusion: the more the data bias, the more the proposed plug-in component improves the performance of the vanilla models. There are two rational explanations for this phenomenon: (1)
Bias-heavy datasets typically have large numbers of contextual representations that potentially induce bias effects. Specifically,
the samples on the EMOTIC and GroupWalk datasets derive from uncontrolled real-world scenarios that contain informative context semantics, such as diverse scene information and agent interaction dynamics. As a result, our component could learn more discriminative context prototypes to serve context-deconfounded training better. Further, causal intervention can more effectively eliminate spurious correlations caused by the adequately extracted confounder and provide sufficient gains.
(2) The implementation of backdoor adjustment~\cite{pearl2009causal} for the causal intervention relies on stratifying the contexts belonging to homogeneous groups by the clustering algorithm. The more severe context bias in the datasets has the bias distribution across more data samples with heterogeneous contexts. In this case, our design can better approximate the theoretical intervention by estimating the average causal effect in stratified contexts that leads to better debiased results.
\textbf{(ii)} Another finding is that CCIM provides richer performance gains for fine-grained methods that capture context semantics. For instance, EmotiCon (average gain of 3.37\% across datasets) with two out-of-subject context modeling branches significantly outperforms EMOT-Net (average gain of 2.22\% across datasets) with only one background context stream on all three datasets. We argue that the essence of fine-grained modeling is the potential context stratification within the sample from the perspective of backdoor adjustment.
Fortunately, CCIM can better refine this stratification effect.

\begin{table*}[t]
\setlength{\tabcolsep}{16pt}
\renewcommand{\arraystretch}{1.1}
\centering
\caption{We show systematic ablation study results on all three datasets. w/ and w/o are short for with and without, respectively.}
\resizebox{\linewidth}{!}{%
\begin{tabular}{c|c|ccc}
\toprule
\multirow{2}{*}{\textbf{Methods}}    & \multirow{2}{*}{\textbf{Different Settings/Designs/Strategies}} & \textbf{EMOTIC}         & \textbf{CAER-S}         & \textbf{GroupWalk}      \\ \cline{3-5} 
                           &                                                & \textbf{mAP} (\%)            & \textbf{Accuracy} (\%)       & \textbf{mAP} (\%)            \\ \midrule
\multirow{12}{*}{EMOT-Net} & Vanilla Model                                  & 27.93          & 74.51          & 58.33          \\
                           & w/ \textbf{CCIM}                                        & \textbf{30.88} & \textbf{75.82} & \textbf{60.74} \\ \cline{2-5} 
                           & w/ Random Dictionary $\bm{Z}$                                    & 26.56          & 73.36         & 57.45          \\
                           & w/ ImageNet Pre-training                       & 28.72          & 74.75          & 58.96          \\ \cline{2-5} 
                           & w/ ResNet-50~\cite{he2016deep}                                   & 29.53          & 75.34          & 59.92          \\
                           & w/ VGG-16~\cite{simonyan2014very}                                      & 28.78          & 74.95          & 59.47          \\ \cline{2-5} 
                           & w/ Additive Attention                          & 30.79          & 75.64          & \textbf{60.85}          \\
                           & w/o $\lambda_{i}$                                          & 30.05          & 75.21          & 59.83          \\
                           & w/o $P(\bm{z}_{i})$                                       & 30.63          & 75.59          & 59.94          \\ \cline{2-5} 
                           & w/o   Masking Strategy                         & 29.86          & 74.84          & 59.22          \\ 
                           & w/  Dichotomous K-Means                       & 30.76          & 75.77          & 60.68          \\
                           & w/  K-Medoids                                 & 30.85          & 75.80          & \textbf{60.77}          \\ \cline{2-5} 
                           & w/ R-FCN~\cite{dai2016r}                       & \textbf{30.88}         & 75.67          & 60.55          \\ 
                            & w/ SSD~\cite{liu2016ssd}                        & \textbf{30.88}          & 75.74          & 60.69          \\
                           \midrule
\multirow{12}{*}{EmotiCon} & Vanilla Model                                  & 35.28          & 88.65          & 65.58          \\
                           & w/ \textbf{CCIM}                                        & \textbf{39.13} & \textbf{91.17} & \textbf{69.31} \\ \cline{2-5} 
                           & w/ Random Dictionary $\bm{Z}$                                    & 35.12          & 87.34          & 65.62          \\
                           & w/ ImageNet Pre-training                       & 37.48          & 90.46          & 68.28          \\ \cline{2-5} 
                           & w/ ResNet-50~\cite{he2016deep}                                   & 38.86          & 90.41          & 68.85          \\
                           & w/ VGG-16~\cite{simonyan2014very}                                      & 37.93          & 89.82          & 68.11          \\ \cline{2-5} 
                           & w/ Additive Attention                          & \textbf{39.16}          & 91.08          & 69.26          \\
                           & w/o $\lambda_{i}$                                          & 38.53          & 89.67          & 68.75          \\
                           & w/o $P(\bm{z}_{i})$                                       & 39.05          & 90.06          & 69.15          \\ \cline{2-5} 
                           & w/o   Masking Strategy                         & 38.06          & 90.57          & 67.79          \\ 
                           & w/   Dichotomous K-Means                       & 39.11          & 91.08          & 69.25          \\
                           & w/   K-Medoids                                 & \textbf{39.16}          & 91.15          & \textbf{69.35}          \\ \cline{2-5} 
                           & w/ R-FCN~\cite{dai2016r}                        & \textbf{39.13}          & 91.04          & 69.16          \\ 
                            & w/ SSD~\cite{liu2016ssd}                        & \textbf{39.13}         & 91.12          & \textbf{69.33}          \\
                           \bottomrule
\end{tabular}
}
\label{ablition}
\end{table*}

\subsection{Ablation Studies}
Table~\ref{ablition} provides systematic ablation studies to evaluate the effectiveness of different settings/designs/strategies when implementing the causal intervention. To investigate the adaptability and necessity of CCIM on different models, we choose the baseline EMOT-Net and SOTA EmotiCon.
The reasons for this are threefold: \textbf{(i)} these two methods are the most representative since they have completely different network architectures and design philosophies; \textbf{(ii)} these two methods significantly differ in the granularities and patterns of modeling context semantics; \textbf{(iii)} there are similar observations and results from other methods in practice.

\subsubsection{Rationality of Confounder Dictionary}
Evaluating the confounder dictionary plays an important role in the causal intervention. 
\textbf{(i)} We first design a random dictionary with identical dimensions to replace the customized dictionary $\bm{Z}$. The random dictionary represents that the confounder dictionary is initialized by random parameterization instead of carefully extracted average context features. We observe that the random dictionary significantly compromises the performance gain of our CCIM. Specifically, the randomized versions of CCIM-based EMOT-Net and EmotiCon decrease their performance by an average of 3.56\% and 3.84\% across the three datasets, respectively.
This observation confirms the effectiveness of our context prototypes and the necessity of the causal implementation.
\textbf{(ii)} Furthermore, we answer what context prototypes are reasonable. To this end, the ResNet-152 pre-trained on the ImageNet dataset~\cite{deng2009imagenet} is employed to extract context features for replacing the default settings regarding the pre-training on the Places-365 dataset. The results are interesting: although the ImageNet-based versions also improve on the vanilla models, they fall short compared to the Places-365-based results. The decreased gains across models suggest that context prototypes based on scene semantics are more conducive to approximating the confounder than those based on object semantics. It is common sense as scene contexts usually include object contexts, \eg, in Figure~\ref{indro1}, ``\textit{grass}'' is the child of the confounder ``\textit{vegetated scenes}''.

 \begin{figure*}[t]
  \centering
  \includegraphics[width=\linewidth]{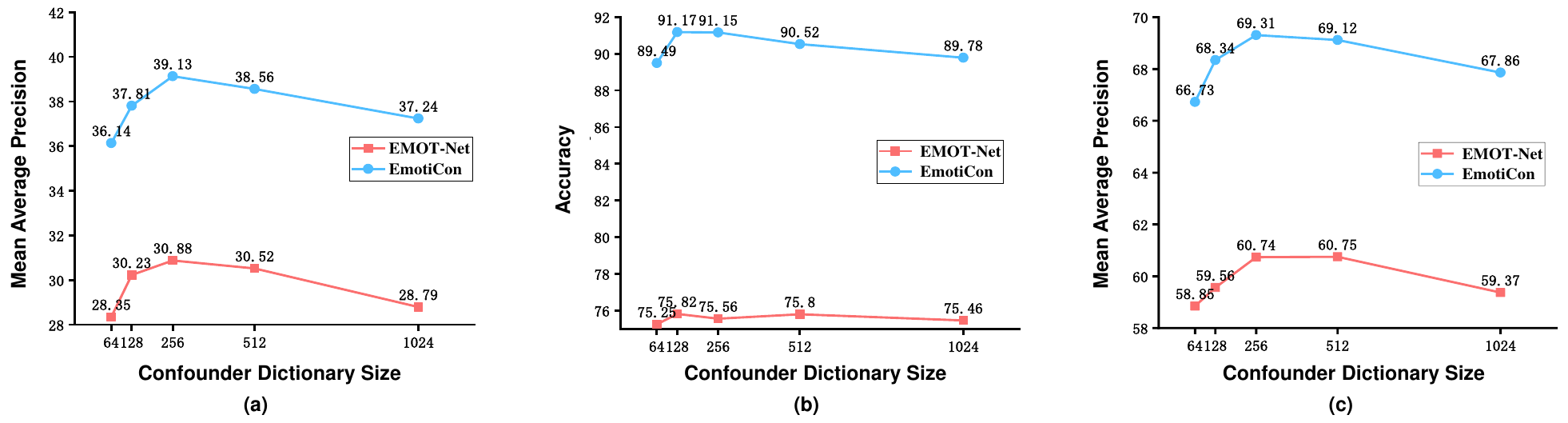}
  \caption{Ablation study results for the size $N$ of the confounder dictionary $\bm{Z}$ on three datasets. (a), (b), and (c) from the EMOTIC, CAER-S, and GroupWalk datasets, respectively.
  }
  \label{size}
\end{figure*}

 \begin{figure*}[t]
  \centering
  \includegraphics[width=\linewidth]{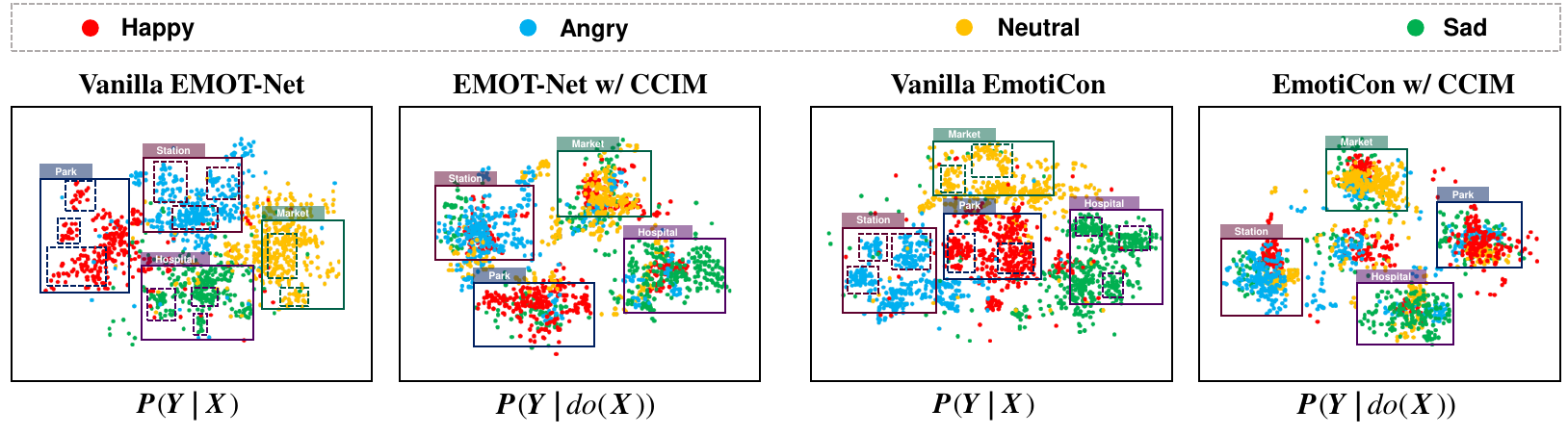}
  \caption{We employ the GroupWalk dataset with four emotion categories to perform the distribution visualization of features for visual clarity. The results from the vanilla and CCIM-based versions of EMOT-Net and EmotiCon are provided to show the differences between $P(\bm{Y}|\bm{X})$ and  $P(\bm{Y}|do(\bm{X}))$.
  }
  \label{vis1}
\end{figure*}

\subsubsection{Robustness of Pre-trained Backbones}
Here, we provide alternative investigations on the pre-trained backbone of extracting the context feature set $M$. The alternatives to the default ResNet-152 are the ResNet-50 and VGG-16 to evaluate the impact on the performance of the same and different families of backbones, respectively. The ablation results from both methods imply that the gains brought by CCIM gradually increase as more advanced pre-trained backbones are introduced. This phenomenon shows that improvements indeed come from CCIM itself rather than depending on a well-chosen pre-trained backbone $\varphi(\cdot)$.

\subsubsection{Effectiveness of Approximate Expectation}
The expectation $\mathbb{E}_{\bm{z}}[g(\bm{z})]$ is the centerpiece for achieving effective causal approximation since it incorporates the extent to which potential confounders $\bm{z}_{i}$ representing distinct context prototypes impact each sample. We perform systematic explorations of different compositions in $\mathbb{E}_{\bm{z}}[g(\bm{z})]$.
\textbf{(i)} First, our proposed additive attention in Equation~(\ref{seven}) is utilized to substitute the default dot product attention for producing the dynamic weight $\lambda_{i}$. The competitive or comparable gains on the three datasets confirm that both attention paradigms are usable and effective, leading to consistent performance improvements.
\textbf{(ii)} Second, when removing the weight $\lambda_{i}$ from the weighted integration process about $\mathbb{E}_{\bm{z}}[g(\bm{z})]$, the decreased results indicate that it is indispensable to characterize the different importance of each confounder.
\textbf{(iii)} Ultimately, we find that considering the prior probability $P(\bm{z}_{i})$ is beneficial in accomplishing causal intervention. It is reasonable because $P(\bm{z}_{i})$ essentially reflects the prior proportion of stratified $\bm{z}_{i}$ on the whole, facilitating a better approximation of the average causal effect by our CCIM.

\subsubsection{Effect of Confounder Size}
The size $N$ of the confounder dictionary reflects the overall confounding degree on a dataset. We set $N$ to 64, 128, 256, 512, and 1024 on all datasets to measure the impact of $N$ on the performance.
As shown in Figure~\ref{size}, when the sizes on the EMOTIC, CAER-S, and GroupWalk datasets are set to 256, 128, and 256, CCIM-based EMOT-Net and EmotiCon achieve the best gains, justifying the default implementation. We conjecture that the smaller confounder size required on the CAER-S dataset is because the samples have limited context scenes and elements as the data are collected in fixed TV shows.
As a result, the vanilla models suffer from the context bias less severely on the CAER-S dataset than the other two datasets.
The above observation suggests that selecting the suitable size $N$ for a dataset containing varying degrees of the harmful bias can well help the models perform de-confounded training.

\subsubsection{Necessity of Masking Strategy}
The masking strategy aims to mask the \textit{recognized subject} to learn prototype representations using pure background contexts. 
The design intuition expects the prototype learning to pay more attention to the background regions that contain more contextual interpretations in a small portion of samples where the recognized subjects occupy a large region.
The gain degradation across all datasets is observed when the target subject regions are not masked. 
The above observation suggests that the masking strategy strengthens our debiasing component, consistently providing valuable improvements for different baselines across all real-world datasets.

\subsubsection{Effect of Clustering Algorithm}
To compute context prototypes, we use the K-Means++ to learn the confounder dictionary $\bm{Z}$. Here, we provide two alternatives (\ie, Dichotomous K-Means and K-Medoids ) to replace the K-Means++ to evaluate the effect on performance.
We observe that the performance difference of the models across all three clustering algorithms is less than 0.13\%, \ie, the choice of clustering algorithm barely affects the performance, demonstrating that the proposed CCIM is robust to the clustering process.

\subsubsection{Effect of Object Detectors}
We also investigate the effect of different object detectors on the confounder dictionary on CAER-S and GroupWalk datasets, which require tagging out the recognized subjects.
Concretely, the default Faster R-CNN in our pipeline is replaced with R-FCN~\cite{dai2016r} and SSD~\cite{liu2016ssd} detectors to perform experiments. 
From the results, Faster R-CNN performs better in most cases, while SSD benefits from the multi-scale feature prediction pattern slightly better than R-FCN. Overall, the effect of different object detectors on the confounder dictionary construction is slight since the gain variation errors for all metrics on both datasets are less than 0.2\%.

 \begin{figure*}[t]
  \centering
  \includegraphics[width=\linewidth]{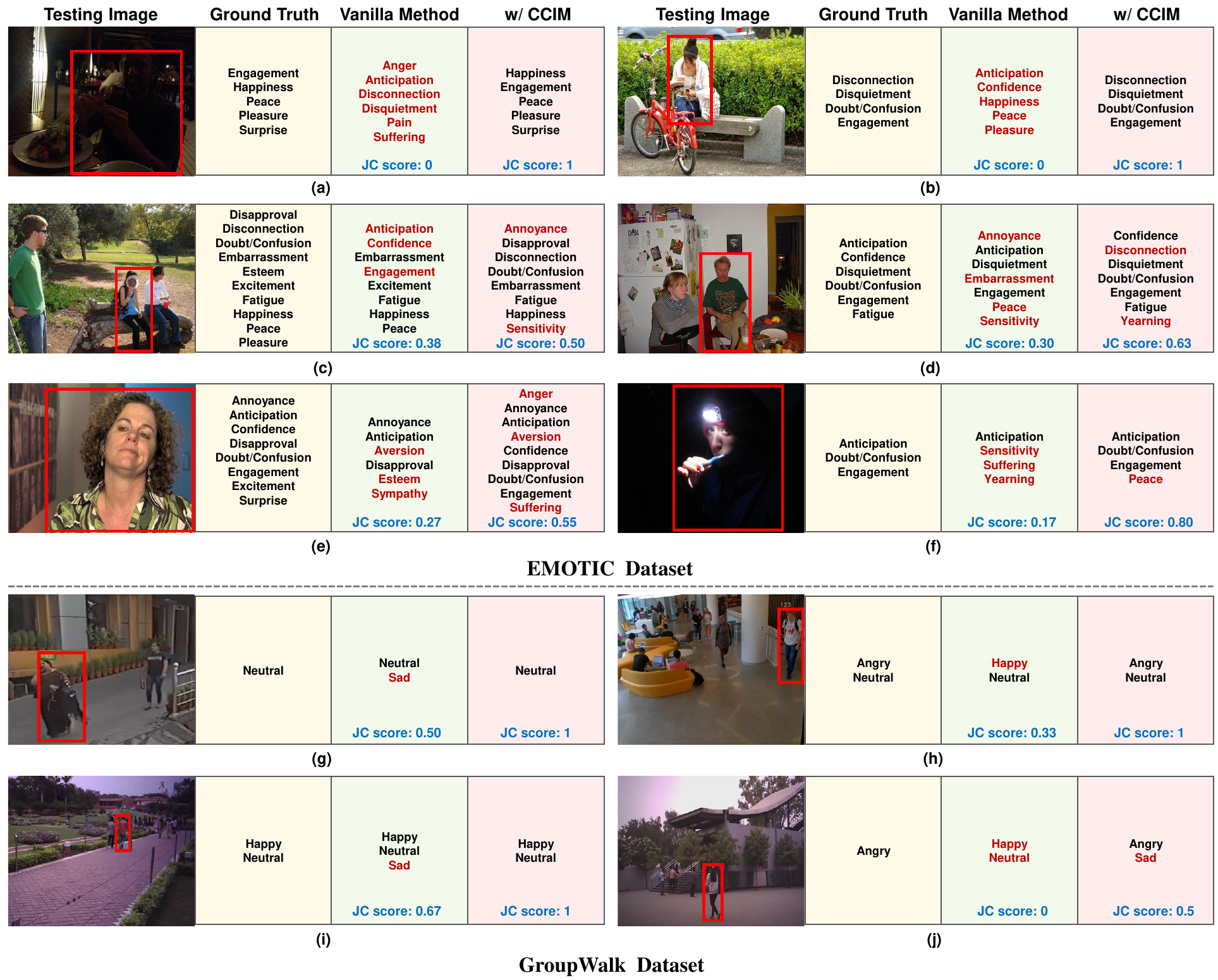}
  \caption{Qualitative results of the vanilla and CCIM-based EmotiCon on the EMOTIC and GroupWalk datasets with different Jaccard coefficient (JC) scores. Incorrectly predicted categories are marked in red.
  }
  \label{vis_emotic_groupwalk}
  \vspace{-5pt}
\end{figure*}

\subsection{Qualitative Evaluation}
\subsubsection{Difference Between Likelihood and Intervention}
Figure~\ref{vis1} visualizes the distributions of context features learned by EMO-Net and EmotiCon on the testing samples to understand the differences between the models
approximate traditional likelihood $P(\bm{Y}|\bm{X})$ and causal intervention $P(\bm{Y}|do(\bm{X}))$. We utilize the GroupWalk dataset due to the modest emotion categories that provide intuitive distinctions visually. Specifically, these sample images contain four types of realistic contexts, \ie, park, market, hospital, and station.
In vanilla models, the context features with the same emotion categories are generally compactly distributed within similar context clusters, \eg, the context features of the hospital with the sad category are closer. This phenomenon implies that context bias causes the models to rely on context-specific spurious correlations for predicting emotions lopsidedly. Conversely, in the CCIM-based models, context-specific features form clusters containing diverse emotion categories.
The distributional change confirms that the causal intervention facilitates the models to fairly integrate each context prototype semantics when predicting emotions, eliminating the detrimental effect of the confounder.

\subsubsection{Qualitative Analysis on the EMOTIC\&GroupWalk}
 \begin{figure}[t]
  \centering
  \includegraphics[width=0.9\linewidth]{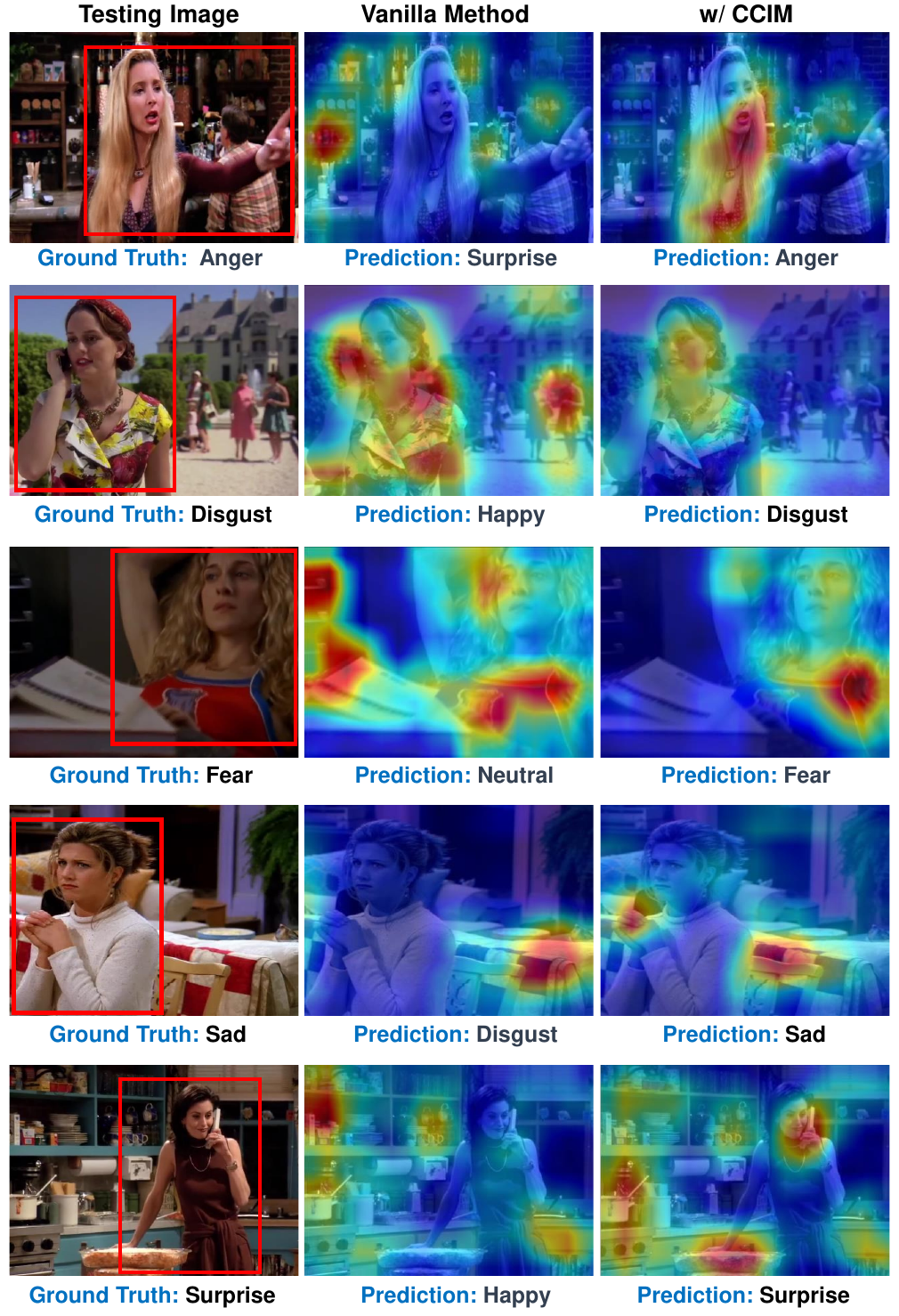}
  \caption{Qualitative results of the vanilla and CCIM-based EMOT-Net on CAER-S dataset. We utilize heat maps to visually capture regions of interest in the model before and after the causal intervention.
  }
  \label{vis_caer}
\end{figure}

As the challenging multi-label classification on the EMOTIC and GroupWalk, we introduce Jaccard Coefficient (JC) scores to more abundantly explain the different performances and roles of our CCIM in different context instances. Despite supporting continuous emotional states on the EMOTIC, we do not use numerical values of VAD dimensions since it is unnatural for humans~\cite{de2023high}.
Concretely, we use the value when $Precision = Recall$ as the threshold for recognizing the category for each discrete emotion. Then, we define the set of predicted emotion categories as $S_p$ and the set of ground truth categories as $S_g$. For each sample, the JC score is calculated as $\left | S_p \cap S_g \right | / \left | S_p \cup S_g \right |$. A higher JC score for a sample means a more accurate prediction by the model, where the maximum value of the score is 1.  Figure~\ref{vis_emotic_groupwalk} shows the performance of the EmotiCon model before and after the intervention. We have the following interesting observations.

\textbf{(i)} On the EMOTIC dataset, EmotiCon is misled by specific contexts to often reason about completely wrong emotions. Taking Figure~\ref{vis_emotic_groupwalk}(a) as an example, since most samples in the training set with contexts related to dim scenes are annotated with negative emotions, the vanilla method gives the opposite predictions to the ground truths for the testing image. Thanks to our causal intervention, CCIM rectifies the context bias and gives the correct categories associated with positive emotions. In addition, the CCIM-based model in Figure~\ref{vis_emotic_groupwalk}(b) decouples the misleadingly positive cues provided by the ``\textit{green vegetatio}'' context in the background, predicting all categories consistent with the ground truths.

\textbf{(ii)} When the subjects' emotional states in background contexts are indistinguishable, the model usually gives ambiguous judgments, leading to poor results. For instance, the JC scores of the original models in Figures~\ref{vis_emotic_groupwalk}(c)\&(d) are only 0.38 and 0.33. Fortunately, CCIM can effectively mitigate the prediction uncertainty and significantly improve the model performance. That is, the JC scores for these two samples improved to 0.50 and 0.63, respectively, 
after the causal intervention.

\textbf{(iii)} We observe similar phenomena on the GroupWalk dataset, \ie, the proposed CCIM consistently improves the performance of the vanilla model in testing samples with diverse contexts. For instance, in Figure~\ref{vis_emotic_groupwalk}(g), CCIM disentangles the spurious correlation between the context (``\textit{hospital entrance}'') and the emotion semantics (``\textit{sad}''), yielding correct results aligned with the ground truth.

 \begin{figure}[t]
  \centering
  \includegraphics[width=1\linewidth]{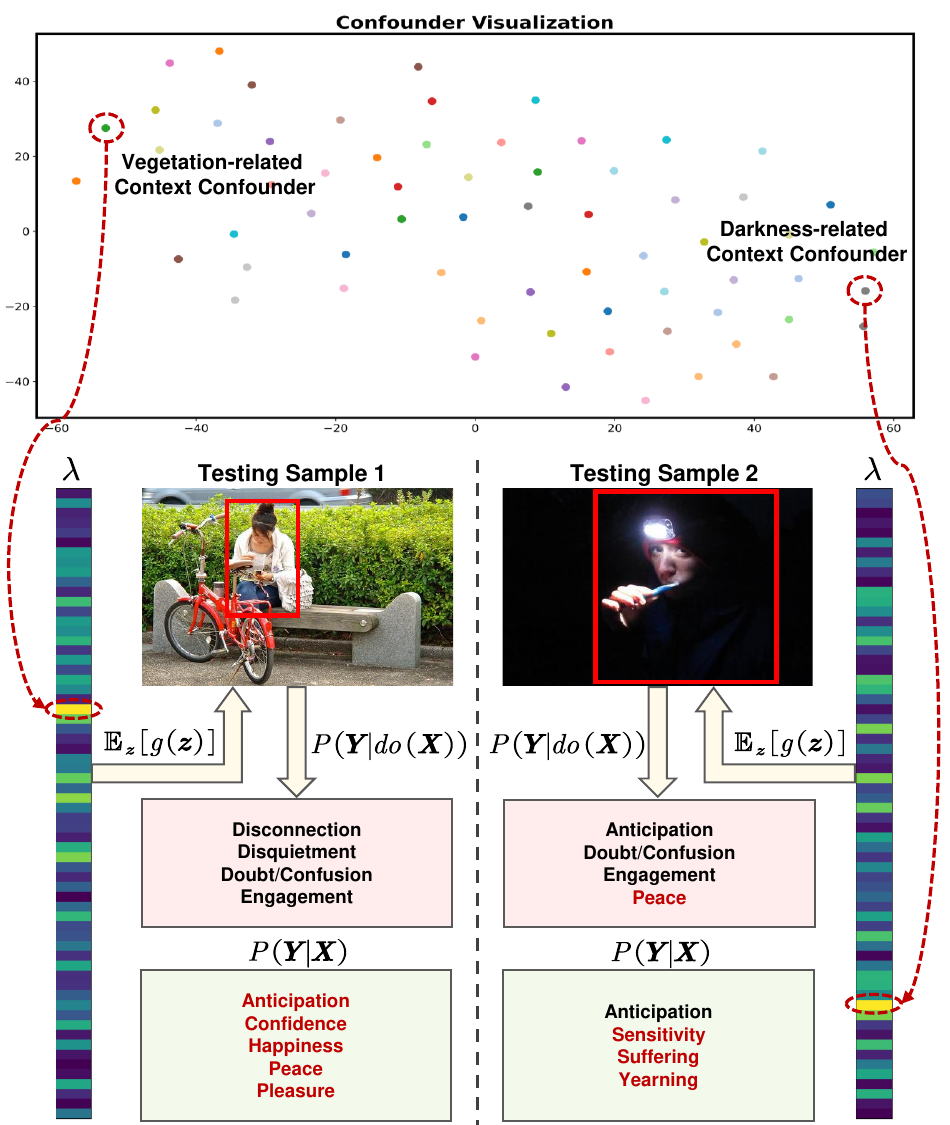}
  \caption{Visualization results of different confounder distributions and the corresponding weight activations in two testing samples. Incorrectly predicted categories are marked in red.
  }
  \label{confounder_vis}
\end{figure}

\subsubsection{Qualitative analysis on the CAER-S}
For the typical multi-class classification in the CAER-S dataset, we randomly show five testing samples with different ground truth emotion categories in Figure~\ref{vis_caer}. Inspired by attention-based efforts~\cite{yang2022emotion,li2021human,li2021sequential}, we use heat maps to observe differences in network models for regions of interest before and after the causal intervention. Here, EMOT-Net is employed instead of EmotiCon to reflect the diversity of evaluation. In heat maps, the red regions imply that the model focuses on during the semantic learning process. The important findings are summarized below.

\textbf{(i)} The vanilla method usually yields undesirable results because of overly gullible beliefs about misleading context cues in the scene. 
In the first row, for example, the context elements around the recognized subject in the baseline's heat map are interpreted to suggest the wrong ``\textit{surprise}'' category. In contrast, CCIM weakens the detrimental effect of contextual stimuli and facilitates the model to pay more attention to anger-related body semantics from the subject region. Similar observations can be found in the fourth and fifth rows of the samples.

\textbf{(ii)} Our component can also correct biased semantics in the subject-focused part in some cases. A typical example is shown in the second row of the sample. By comparing the heat maps before and after applying the CCIM, the causal intervention helps the model to capture the facial prompts from the subject reflecting the ``\textit{disgust}'' emotion, resulting in the correct prediction. Interestingly, our component simultaneously removes context cues for agent interactions in the right background that may cause the ''\textit{happy-context}'' mapping prejudice. Similar capabilities are recognized in the third row of the sample.

\subsubsection{Confounder Visualization}
To intuitively understand the confounder impact in causal intervention, we visualize 64 clustering centers representing different context confounder prototypes. From Figure~\ref{confounder_vis}, distinct prototypes are well separated distributionally, verifying that our strategy can correctly model stratified confounder features.
Then, we perform testing experiments on the EMOTIC dataset using the CCIM-based EmotiCon as the baseline.
While inferring two testing samples, we visualize weight maps of confounders of the corresponding coefficient set $\lambda$ in intervention $P(\bm{Y}|do(\bm{X}))$, where brighter colors represent higher values.
In Sample 1, the corresponding weight of the vegetation-related confounder is activated higher to facilitate our component to disentangle the spurious correlation between vegetation context and positive sentiment, leading to reasonable predictions.
In Sample 2, our causal intervention gives greater attention to the darkness-related confounder and forces the model to correct the erroneously negative emotions induced by the dark scene in the vanilla $P(\bm{Y}|\bm{X})$.
In summary, CCIM can dynamically decouple the effects of the context bias to different degrees for samples, enabling the model to extract meaningful cues related to the correct emotions in the de-confounded training.

\section{Conclusion and Discussion}
\label{con}
This paper proposes a causal debiasing component to reduce the harmful bias of uneven distribution of emotional states across diverse contexts in the CAER task. As a first in-depth investigation of the context, 
we disentangle the causalities among variables via a tailored causal graph and present a Contextual Causal Intervention Module (CCIM) to remove the adverse effect caused by the context bias as a confounder.
Systematic experiments demonstrate the reasonableness of the causality-driven learning paradigm. It is worth noting that our causal weapon can be applied to other context-aware tasks to facilitate the progress of the community.

This work has potential applications and broad impacts in other fields.
\textbf{(i)} CCIM can be readily extended to other context-driven tasks to promote unbiased estimation in the corresponding domains, such as egocentric action anticipation and salient object detection.
Through causal debiasing, our component can help researchers build task-specific context confounders and breakthrough performance bottlenecks in vanilla models.
\textbf{(ii)} The proposed intervention paradigm can be extended to temporal context scenarios to facilitate biased interference due to temporal asynchrony in sequential modeling applications. Specifically, the context stratification strategy can capture the average causal effect in long-range contextual dependencies from temporal representations to boost de-confounded training.
\textbf{(iii)} This work contributes to improving the fairness of baseline methods in context-aware tasks and preventing potential discrimination due to bias in deep models. Related techniques offer promising solutions for developing trustworthy intelligent systems.

% \ifCLASSOPTIONcompsoc
%   % The Computer Society usually uses the plural form
%   \section*{Acknowledgments}
% \else
%   % regular IEEE prefers the singular form
% \fi

% This work is supported in part by the National Key R\&D Program of China (2021ZD0113503) and in part by the Shanghai Municipal Science and Technology Major Project (2021SHZDZX0103).

\ifCLASSOPTIONcaptionsoff
  \newpage
\fi



\bibliographystyle{IEEEtran}
\bibliography{main}

\vspace{-25pt}
\begin{IEEEbiography}[{\includegraphics[width= 1in,height=1.3in,clip,keepaspectratio]{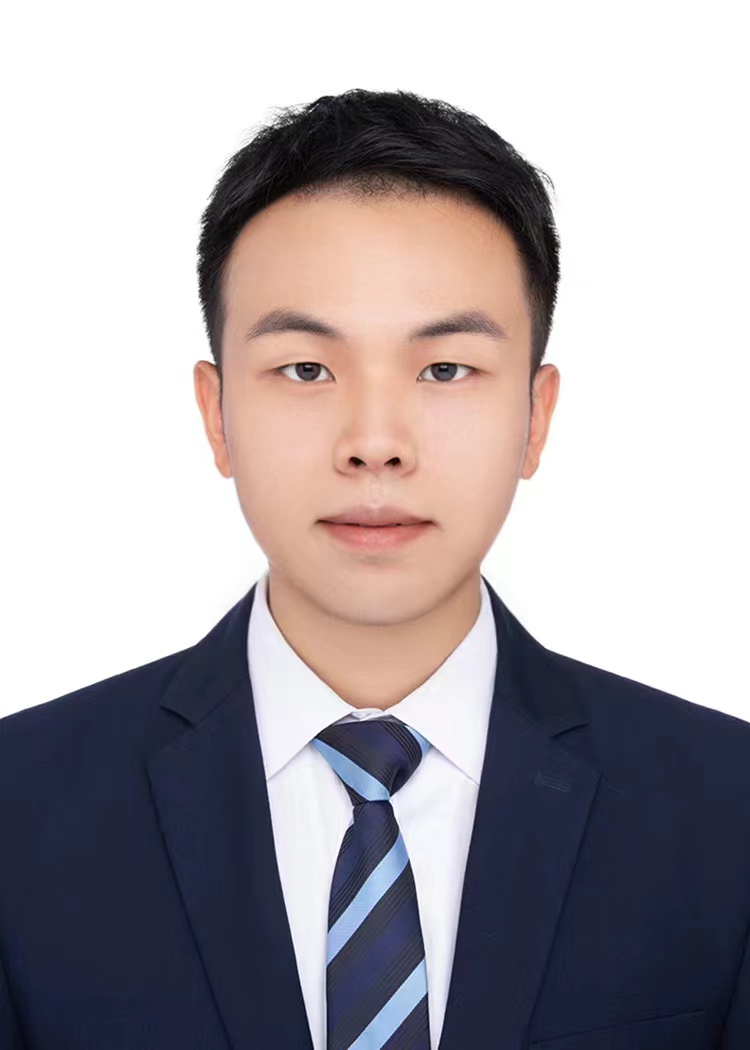}}]
{Dingkang Yang} received the B.E. degree in Communication Engineering from the joint training of Yunnan University and the Chinese People's Armed Police (PAP), Kunming, China, in 2020. He is currently pursuing the Ph.D. degree at the Academy for Engineering and Technology, Fudan University, Shanghai, China. His research interests include multimodal learning, affective computing, causal inference, and large language models.
\end{IEEEbiography}

\vspace{-25pt}
\begin{IEEEbiography}[{\includegraphics[width=1in,height=1.3in,clip,keepaspectratio]{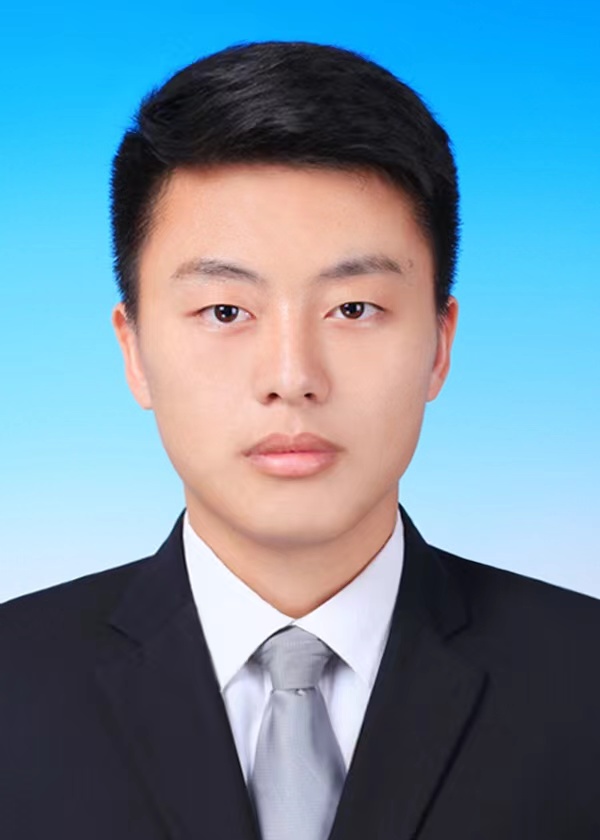}}]
{Kun Yang} received the B.S. degree in Automation from the Donghua University, Shanghai, China, in 2020. He is currently pursuing the Ph.D. degree at the Academy for Engineering and Technology, Fudan University, Shanghai, China. His research interests include collaborative perception, autonomous vehicles, and vehicular edge computing. He is currently working at the Fudan Institute on Networking Systems of AI, Fudan University.
\end{IEEEbiography}

\vspace{-25pt}
\begin{IEEEbiography}[{\includegraphics[width=1in,height=1.3in,clip,keepaspectratio]{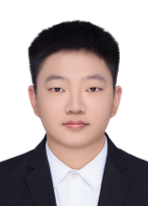}}]
{Haopeng Kuang} received the M.Sc. degree in Mathematics from the College of Mathematics, Jilin University, Changchun, China, in 2019. He is currently pursuing the Ph.D. degree at the Academy for Engineering and Technology, Fudan University, Shanghai, China. His research interests include multimodal learning and the applications of artificial intelligence in healthcare.
\end{IEEEbiography}

\vspace{-25pt}
\begin{IEEEbiography}[{\includegraphics[width=1in,height=1.3in,clip,keepaspectratio]{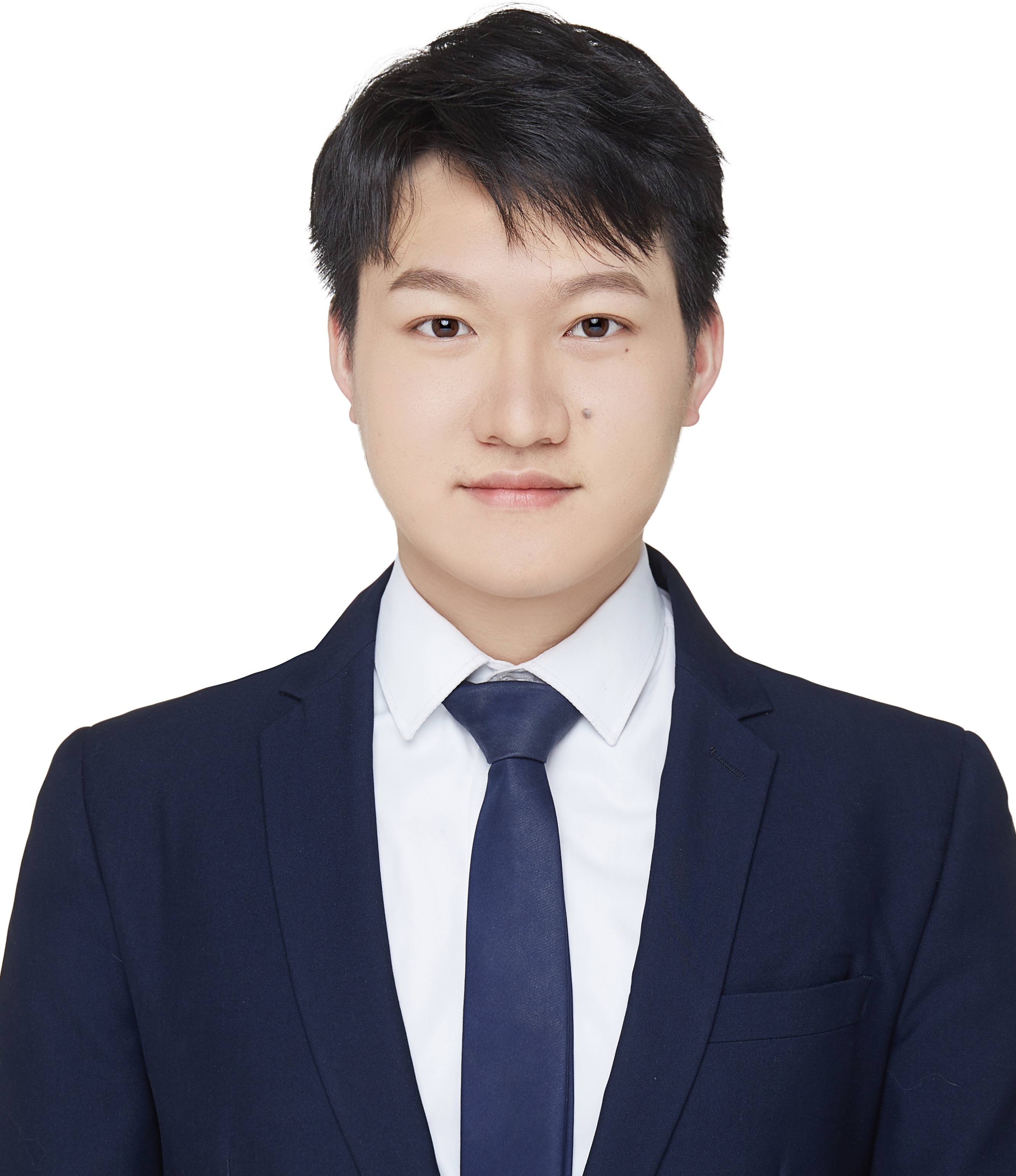}}]
{Zhaoyu Chen} received the B.E. degree from Shandong University by 2020. He is currently pursuing the Ph.D. degree with the Academy for Engineering and Technology, Fudan University, Shanghai. His research interests include artificial intelligence security, computer vision, and their applications, such as adversarial examples and semantic segmentation.
\end{IEEEbiography}

\vspace{-25pt}
\begin{IEEEbiography}[{\includegraphics[width=1in,height=1.3in,clip,keepaspectratio]{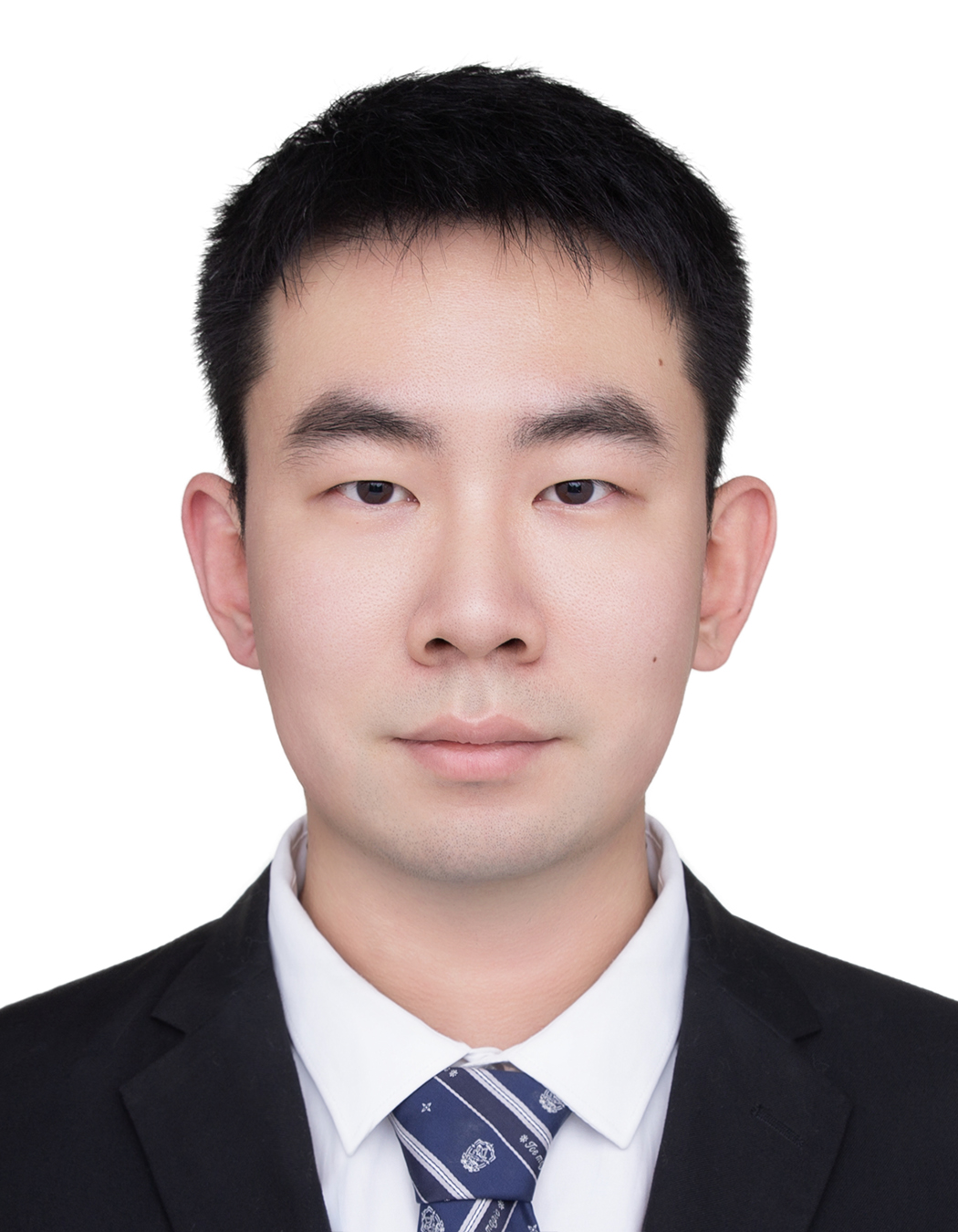}}]
{Yuzheng Wang} received the B.E. degree in intelligent science and technology from
Nankai University, Tianjin, China, in 2020. He is working as a Ph.D student at the Academy for Engineering and Technology, Fudan University, Shanghai, China. His research interests include knowledge distillation, adversarial robustness, novel class discovery, etc. The main purpose is to enhance the mobile deployment of deep learning models, overcome the unavailability of private data at the user end, improve the robustness of the model, and reduce the risk of malicious attacks.
\end{IEEEbiography}

\vspace{-25pt}
\begin{IEEEbiography}[{\includegraphics[width=1in,height=1.3in,clip,keepaspectratio]{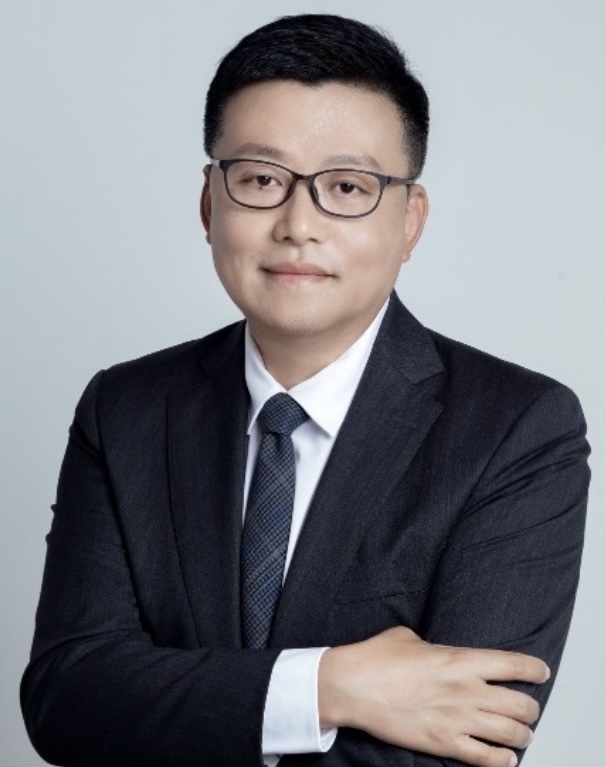}}]
{Lihua Zhang} received the Ph.D. degree from the Department of Automation, Tsinghua University, Beijing, China, in 2000. He is currently a Professor at the Academy for Engineering and Technology, Fudan University, Shanghai, China. In recent years, he has participated in a number of national science and technology research and development projects as a project leader and sub-project. His current research interests are in artificial intelligence and its applications, including machine intuition, computer vision and intelligent perception, virtual reality and digital twinning, intelligent robotics and unmanned systems, intelligent computing and intelligent chips, intelligent healthcare, intelligent connected vehicles, etc.
\end{IEEEbiography}

\end{document}